\documentclass[twoside]{article}

\usepackage{fancyhdr}
\newlength\titlebox 
\usepackage[round]{natbib}

\usepackage{url}            
\usepackage{booktabs}       
\usepackage{amsfonts}       
\usepackage{nicefrac}       
\usepackage{microtype}      
\usepackage{xcolor}         

\usepackage{bbm}

\usepackage{color, colortbl}
\definecolor{win}{rgb}{1,1,0}
\usepackage{stfloats}
\usepackage{xcolor}
\usepackage[noend]{algpseudocode}
\usepackage{adjustbox}
\usepackage{enumerate}
\usepackage{array}
\newcommand{\PreserveBackslash}[1]{\let\temp=\\#1\let\\=\temp}
\newcolumntype{C}[1]{>{\PreserveBackslash\centering}p{#1}}
\usepackage{etoolbox}
\usepackage{comment}

\usepackage{amsmath, amsthm}
\theoremstyle{plain}
\newtheorem{thm}{Theorem}
\newtheorem{lem}[thm]{Lemma}

\newtheorem{assum}[thm]{Assumption}
\theoremstyle{definition}

\theoremstyle{remark}

\usepackage{amsfonts}
\usepackage{latexsym}
\usepackage{amssymb,amsbsy}
\usepackage{url}
\usepackage{mathtools}
\usepackage{amscd}

\usepackage{xcolor,stackengine}

\usepackage{upref}
\usepackage{graphicx}
\usepackage{psfrag}
\usepackage{multirow}
\usepackage{mathtools}
\usepackage{xparse}
\usepackage{bbm}
\usepackage{subcaption}
\usepackage{hyperref}
\definecolor{darkgreen}{rgb}{0,0.5,0}
\definecolor{darkblue}{rgb}{0,0,0.5}
\hypersetup{
    colorlinks=true,
    linkcolor=darkblue,
    citecolor=darkgreen,
    filecolor=magenta,
    urlcolor=blue
}
\usepackage[linesnumbered,ruled,vlined]{algorithm2e}

\DeclarePairedDelimiterX{\set}[1]{\{}{\}}{\setargs{#1}}
\NewDocumentCommand{\setargs}{>{\SplitArgument{1}{;}}m}
{\setargsaux#1}
\NewDocumentCommand{\setargsaux}{mm}
{\IfNoValueTF{#2}{#1} {#1\,\delimsize|\,\mathopen{}#2}}

\DeclarePairedDelimiter\abs{\lvert}{\rvert}

\DeclareMathOperator*{\Do}{do}
\DeclareMathOperator*{\diag}{\bf{diag}}

\usepackage{tikz}
\usetikzlibrary{positioning}

\newcommand{\R}{\mathbb{R}}
\newcommand{\indep}{\perp \!\!\! \perp}
\newcommand{\E}{\mathbb{E}}
\newcommand{\given}{\mid}
\newcommand{\tvec}[1]{\hat{\vec{#1}}}
\newcommand{\ind}{\mathbbm{1}}
\renewcommand\vec{\mathbf}

\newcommand{\ddelta}{\boldsymbol{\delta}}
\newcommand{\aalpha}{\boldsymbol{\alpha}}
\newcommand{\bbeta}{\boldsymbol{\beta}}
\newcommand{\ggamma}{\boldsymbol{\gamma}}
\newcommand{\norm}[1]{\left\lVert#1\right\rVert}

\newcommand{\kron}{\mathbbm{1}}
\newcommand{\vPhi}{\vec{\Phi}}
\newcommand{\vTheta}{\vec{\Theta}}
\newcommand{\vA}{\vec{A}}
\newcommand{\vX}{\vec{X}}
\newcommand{\vZ}{\vec{Z}}
\newcommand{\ModelA}{\textbf{\textit{Model A}}}
\newcommand{\ModelB}{\textbf{\textit{Model B}}}
\newcommand{\ModelC}{\textbf{\textit{Model C}}}
\newcommand{\strengthZT}{{\textcolor{blue}{\mu_{zt}}}}
\newcommand{\strengthXY}{{\textcolor{red}{\mu_{xy}}}}

\begin{document}

%

%

\twocolumn

\title{Synthetic Potential Outcomes and Causal Mixture Identifiability}

\author{%
  Bijan Mazaheri\thanks{BM is supported by a postdoctoral fellowship at the Eric and Wendy Schmidt Center at the Broad Institute of MIT and Harvard. \url{http://bijanmazaheri.com}} \\
  Eric and Wendy Schmidt Center\\
  Broad Institute of MIT and Harvard\\
  Cambridge, MA 02142 \\
  \texttt{bmazaher@broadinstitute.org} \\
   \and
  Chandler Squires\thanks{CS was partially supported by ONR (N00014-22-1-2116) and DOE-ASCR (DE-SC0023187). 
}\\
  Laboratory for Information and Decision Systems\\
  Eric and Wendy Schmidt Center\\
  Broad Institute of MIT and Harvard\\
  Cambridge, MA\\
  \texttt{csquires@mit.edu}\\
   \and
    Caroline Uhler\thanks{CU was partially supported by NCCIH/NIH (1DP2AT012345), ONR (N00014-22-1-2116), DOE-ASCR (DE-SC0023187), the MIT-IBM Watson AI Lab, the Eric and Wendy Schmidt Center at the Broad Institute, and a Simons Investigator Award.}\\
  Laboratory for Information and Decision Systems\\
  Eric and Wendy Schmidt Center\\
  Broad Institute of MIT and Harvard\\
  Cambridge, MA\\
  \texttt{cuhler@mit.ed}
}
\maketitle
\begin{abstract}

Heterogeneous data from multiple populations, sub-groups, or sources is often represented as a “mixture model” with a single latent class influencing all of the observed covariates. Heterogeneity can be resolved at multiple levels by grouping populations according to different notions of similarity. This paper proposes grouping with respect to the causal response of an intervention or perturbation on the system. This definition is distinct from previous notions, such as similar covariate values (e.g. clustering) or similar correlations between covariates (e.g. Gaussian mixture models). To solve the problem, we ``synthetically sample'' from a counterfactual distribution using higher-order multi-linear moments of the observable data. To understand how these ``causal mixtures'' fit in with more classical notions, we develop a hierarchy of mixture identifiability.
\end{abstract}

\section{INTRODUCTION}

\subsection{Causal Inference}
Causality encompasses both counterfactual (what could have been) and hypothetical (what could be) statements. If a patient is given treatment and cured, saying that treatment \emph{caused} recovery is equivalent to saying that the patient would not have recovered without treatment (a counterfactual). Similarly, recommending a treatment carries with it the implication that the patient is better off with treatment than without.

\paragraph{Latency of Counterfactuals}
When hoping to recover causal effects, we are limited to data from a single world, revealing only one ``potential outcome'' (e.g. the patient recovers under treatment). As such, the counterfactual remains unobserved --- for all we know, the treatment had no effect and the patient would have recovered either way. If a treated patient's recovery is $Y^{(1)}$ and the untreated counterfactual is $Y^{(0)}$, then the individual treatment effect (ITE) is the difference between these ``potential outcomes'': $Y^{(1)} - Y^{(0)}$.
The inherent latency of counterfactual and hypothetical outcomes is referred to as the ``fundamental problem of causal inference'' \citep{imbens2015causal}.

\paragraph{Exchangeability}
In order to estimate causal effects, we must ``pair'' examples deemed ``exchangeable.'' For example, two twins given different treatments may be considered to be approximate counterfactuals of each other. When twins are not available, a common approach is to instead pair \emph{populations} to identify an ``average treatment effect'' (ATE) $\E(Y^{(1)} - Y^{(0)})$. This can be achieved with a randomized control trial (RCT), which randomly assigns ``treatment'' and ``control'' groups to ensure exchangeability in expectation.

RCTs are not always feasible --- it is unethical to withhold potentially lifesaving medicine from sick patients, and many economic policies must be implemented without a preliminary test. In such settings, methods for causal inference have developed across a number of fields \citep{pearl2009causality, imbens2015causal, peters2017elements}.

\paragraph{Refining Exchangeability}
Approaches to causal inference all rely on the insight that two individuals (or groups) need not be identical in every way to provide access to a counterfactual. Epidemiological methods often make use of assumptions like ``unconfoundedness'' (also called strong ignorability), which is formalized as $Y^{(1)},Y^{(0)} \indep T \given \vec{X}$ for treatment $T$ and covariates $\vec{X}$ \citep{rosenbaum1983central}. This assumption implies that no unobserved $U$ confounds both $T$ and $Y$, implicitly guaranteeing that two data entries with identical $\vec{X} = \vec{x}$ and different $T=t$ are sufficiently exchangeable to be considered counterfactuals.

Relaxing unconfoundedness, graphical methods can be used to model the relationships between covariates \citep{lauritzen1996graphical, koller2009probabilistic}. Such models define ``adjustment sets'' of covariates that are sufficient for exchangeability \citep{pearl2009causality}. These approaches allow for exchangeability only with respect to a subset of $\vec{X}$ --- something that is needed if $\vec{X}$ contains a common effect of $T, Y$ (e.g. $T \rightarrow X \leftarrow Y$). However, they still cannot handle unobserved confounding due to $U$ unless a valid adjustment set can fully resolve its effect on the ATE.

\paragraph{Identifiability}
Tools for causal inference are often employed when there is no ground truth validation for success. Therefore it is essential to guarantee identifiability of the desired quantity, i.e. that infinite data from a distribution will uniquely correspond to a single answer. The work of Judea Pearl has heavily explored the identification of ATEs using adjustment sets that are given with respect to graphically modeled unobserved confounding \citep{pearl2009causality}. This approach links causal identifiability to the existence of an observable metric of exchangeability.

\subsection{Latent Heterogeneity} \label{sec: heterogeneity}

The principal culprit for entanglement between causation and correlation is latent heterogeneity. In the absence of an observable adjustment set, it is impossible to pair exchangeable points or distributions. 
For example, if groups with differing rates of recovery are also given treatment at different rates, then the apparent treatment effect will be biased. Treatment is often given to the most severe cases of a disease, resulting in treated patients being associated with poorer outcomes despite a positive treatment response. This is known as Simpson's Paradox \citep{simpson1951interpretation}.

\paragraph{Formal Setting}
To study this setting rigorously, we consider a $k$-mixture to be given by:
\begin{itemize}\setlength\itemsep{0em}
    \item A binary treatment or action $T \in \{0, 1\}$.
    \item An outcome $Y$ (discrete or continuous).
    \item A list of covariates which we will simplify to $\vec{X}, \vec{Z}$ (discrete or continuous), which can be scalar or vector-valued so long as certain identifiability constraints are met.
    \item $k$ classes given by $U\in [k]$.
\end{itemize}
We will generally assume that $U$ has nontrivial causal arrows to all four $\{T, Y, \vec{X}, \vec{Z}\}$. 
These ``nontrivial arrows'' imply an ``overlap'' or ``positivity'' assumption of $0 < \Pr(T \given U) < 1$. We restrict ourselves to an ``unconfounded'' setting where $Y^{t} \indep T \given U$ for simplicity while noting that extending to adjustment sets of the form $\vec{C} \cup \{U\}$ for observable $\vec{C}$ is straightforward.
In addition, we will assume $\vec{X}, \vec{Z}$ can be transformed into feature maps $\vPhi(\vec X), \vTheta(\vec Z) \in \mathbb{R}^k$ with distinct expectations within each $U$. That is $\E[\Phi \given u] \neq \E[\Phi \given u']$ whenever $u \neq u'$ for all $\Phi \in \vPhi(\vec X)$ (and similarly for $\vTheta$). To simplify the presentation of our algorithm, we will assume $\vec{X}, \vec{Z} \in \mathbb{R}^k$ instead of directly referring to their feature maps. The requirement of ``distinctness'' manifests as the invertibility of a matrix used in our algorithm, which is empirically verifiable.

\subsection{Mixtures of Treatment Effects}
We define the ``mixture of treatment effects'' (MTEs) problem to be the task of uniquely identifying conditional average treatment effects on a latent class $U$, e.g. $\E(Y^{(1)} - Y^{(0)} \given u)$, in addition to the probability of each class $\Pr(u)$.

To illustrate the importance of identifying causal heterogeneity, consider two possible populations of a bacterial infection. One consists of a sub-population that is resistant to an antibiotic, as well as a non-resistant sub-population. Another bacterial population is homogeneously partially resistant to the antibiotic, requiring a higher dosage to be killed. In the first setting, use of the antibiotic is ill-advised, as it will allow the resistant subpopulation to dominate. In the second setting, a larger dose is appropriate.

A similar problem emerges within virus vaccination. A vaccine may lose average effectiveness over time due to a loss in antibodies (a homogeneous loss in potency) or the emergence of a new resistant variant (a heterogeneous response). In this setting, the homogeneous response should be combated with a booster, while the heterogeneous response requires a new vaccine.

A final motivating task is that of evaluating the success of an intervention's implementation. For example, imperfect participation or a faulty shipment may create a heterogeneous response to a treatment, while general ineffectiveness may be homogeneous.

\subsection{Identifiability Hierarchy}\label{sec: identifiability hierarchy}

There are multiple notions of identifiability within causally heterogeneous settings. The first is that of the heterogeneous treatment effects (HTEs) with respect to a subset $\vec{A}'$ of adjustment set $\vec{A}$, which is only identifiable if $\vec{A}'$ is observable. In principal, this means that \emph{each entry} in a dataset can grouped according to its conditional average treatment effect (CATE) indexed by a set of values for $\vec{A}'$. In our setup, $\vec{A}' = \vec{A} = \{U\}$, and $U \in \vec{A}$ regardless of the causal structure on $(\vec Z, T, Y, \vec X)$.

In contrast to the \emph{endogenous} and \emph{directly observed} heterogeneity of HTEs, the causal effects within MTEs are \emph{exogenously} and \emph{latently} heterogeneous. The primary computational challenge when recovering MTEs is the latency of $U$, which disallows the direct observation of CATEs with respect to $U$. Mixture models are considered identifiable when observed statistics map uniquely to the parameters of the model, i.e. $\Pr(U), \Pr(T, Y, \vec X, \vec Z \given u)$, up to the $k!$ symmetric permutations of $U$. If a mixture model is identified, MTEs can be recovered with respect to adjustment sets including unobserved $U$, even though such treatment effects cannot be assigned to individual entries of data. 

At an even coarser resolution than MTEs, ATEs are considered identifiable when observed statistics map uniquely to the average causal effects, e.g. $\E(Y^{(1)} - Y^{(0)})$, which marginalizes out the effect of $U$.

These notions of identifiability are hierarchical. Clearly, datapoints that have been grouped by HTE can be separately analyzed to recover the CATEs needed for MTEs. Furthermore, MTEs can be marginalized into an ATE using the law of total probability. Meanwhile, negative controls \citep{miao2018confounding} identify ATEs without giving access to MTEs or HTEs, and mixture models can identify MTEs without giving access to point-specific mixture membership. It is essential to develop methods at the level of the scientific goal --- using a method that identifies a ``more coarse'' level can lead to uncertainty in results and using methods that are ``more fine'' have more restrictive identifiability requirements.

Particular interest has been given to mixture model methods as tools for causal inference because they give access to the joint probability distribution between $U$ and $T,Y$, allowing the identification of both ATE and MTEs. However, two mixture models can easily generate identical ATEs. While less obvious, two mixture models can also generate identical MTEs, motivating an additional level of identifiability.

\subsection{Contributions}
Mixture model identifiability does not precisely correspond to MTE identifiability. That is, two different distributions for $\Pr(\vec Z, T, Y, X \given u)$ could map to the same MTEs. In particular, correctly identifying the distribution on the proxies $X, Z$ is inconsequential for recovering the conditional average treatment effects with respect to $U$. For this reason, the conditions for MTE identifiability are \emph{milder} than those for mixture identification. We present a straightforward differentiation between these conditions as a hierarchy on four levels:
\begin{enumerate}\setlength\itemsep{0em}
    \item Identification of HTEs, requiring observed $U$.
    \item Identification of the mixture components, requiring $\vec{Z} \indep (T,Y) \indep \vec{X} \given U$.
    \item Identification of MTEs, requiring $\vec{Z} \indep Y \given T, U$ and $\vec{X} \indep (Y, T) \given U$.
    \item Identification of ATEs, requiring $\vec{Z} \indep Y \given T, U$ and $\vec{X} \indep T \given U$.
\end{enumerate}
The four notions of identifiability require decreasingly restrictive assumptions, which can be interpreted graphically as adding possible edges to the $5$-vertex graph shown in Figure~\ref{fig: requirements and hierarchy}. The full requirements for identifiability at levels 3 and 4 are formally given in Theorems~\ref{thm: ATE identifiability} and \ref{thm: MTE identifiability} after introducing some notation \footnote{The independence conditions needed for identification can be relaxed into a notion of exchangeability in expectation, since all that is required is for expectations to factorize. This will be discussed in more detail in Section~\ref{sec: discussion}.}.

\begin{figure}
    \centering
    \scalebox{1}{\begin{tikzpicture}
\draw[rounded corners=10pt] (-.45,-.2) rectangle (2.45,1.5);
\node[] at (1, 1.2) {1. HTE};
\node at (1, .4) {\scalebox{.35}{\begin {tikzpicture}[-latex ,auto ,node distance =2 cm and 2 cm ,on grid , ultra thick, state/.style ={circle, draw, minimum width =1 cm, ultra thick}, cstate/.style ={ circle, draw, minimum width =.5 cm, fill=black, text = white}]
        \node[state] (T) {$T$};
        \node[state] (Y)[right =of T] {$Y$};
        \node[state] (U)[above left = 2cm and 1cm of Y] {$U$};
        \path (U) edge[] (T) (U) edge[] (Y) (T) edge (Y);
\end{tikzpicture}}};
\draw[rounded corners=10pt] (-0.55,-1.9) rectangle (2.55,1.6);
\node[] at (1, -.5) {2. Full Mixture};
\node at (1, -1.3) {\scalebox{.35}{\begin {tikzpicture}[-latex ,auto ,node distance =2 cm and 2 cm ,on grid , ultra thick, state/.style ={circle, draw, minimum width =1 cm, ultra thick}, cstate/.style ={ circle, draw, minimum width =.5 cm, fill=black, text = white}]
        \node[state] (T) {$T$};
        \node[state] (Y)[right =of T] {$Y$};
        \node[state] (X)[right =of Y] {$\vec X$};
        \node[state] (Z)[left =of T] {$\vec Z$};
        \node[state, dashed, green!50!black] (U)[above left = 2cm and 1cm of Y] {$U$};
        \path (U) edge[dashed] (T) (U) edge[dashed] (Y) (T) edge (Y) (U) edge[dashed] (X) (U) edge[dashed] (Z);
\end{tikzpicture}}};
\draw[rounded corners=10pt] (-0.65,-3.6) rectangle (2.65,1.7);
\node[] at (1, -2.2) {3. MTE};
\node at (1, -3) {\scalebox{.35}{\begin {tikzpicture}[-latex ,auto ,node distance =2 cm and 2 cm ,on grid , ultra thick, state/.style ={circle, draw, minimum width =1 cm, ultra thick}, cstate/.style ={ circle, draw, minimum width =.5 cm, fill=black, text = white}]
        \node[state] (T) {$T$};
        \node[state] (Y)[right =of T] {$Y$};
        \node[state] (X)[right =of Y] {$\vec X$};
        \node[state] (Z)[left =of T] {$\vec Z$};
        \node[state, dashed, green!50!black] (U)[above left = 2cm and 1cm of Y] {$U$};
        \path (U) edge[dashed] (T) (U) edge[dashed] (Y) (T) edge (Y) (U) edge[dashed] (X) (U) edge[dashed] (Z);
        \path[color=blue] (Z) edge[<->] (T);
\end{tikzpicture}}};
\draw[rounded corners=10pt] (-0.75,-5.3) rectangle (2.75,1.8);
\node[] at (1, -3.9) {4. ATE};
\node at (1, -4.7) {\scalebox{.35}{\begin {tikzpicture}[-latex ,auto ,node distance =2 cm and 2 cm ,on grid , ultra thick, state/.style ={circle, draw, minimum width =1 cm, ultra thick}, cstate/.style ={ circle, draw, minimum width =.5 cm, fill=black, text = white}]
        \node[state] (T) {$T$};
        \node[state] (Y)[right =of T] {$Y$};
        \node[state] (X)[right =of Y] {$\vec X$};
        \node[state] (Z)[left =of T] {$\vec Z$};
        \node[state, dashed, green!50!black] (U)[above left = 2cm and 1cm of Y] {$U$};
        \path (U) edge[dashed] (T) (U) edge[dashed] (Y) (T) edge (Y) (U) edge[dashed] (X) (U) edge[dashed] (Z);
        \path[color=blue] (Z) edge[<->] (T);
        \path[color=red] (X) edge (Y);
\end{tikzpicture}}};
\end{tikzpicture}}
    \caption{HTEs are only identifiable if $U$ is observed. Identification of the next three levels require decreasingly restrictive graphical assumptions, demonstrated by the addition of an edge. $Z {\color{blue} \leftrightarrow} T$ indicates an arrow that could go either direction (or a bidirected arrow from unobserved confounding).}
    \label{fig: requirements and hierarchy}
\end{figure}

Identification at levels 2 and 4 are addressed for categorical (discrete) $X, Z$ in previous works discussed in Section~\ref{sec:related works}, which we expand to continous $X, Z$ as a secondary result. Our primary contribution is to give the first identifiability result for level 3, including an algorithm called ``synthetic potential outcomes'' (SPOs) that we develop in Section~\ref{sec: SPOs}. Section~\ref{sec: hierarchy} then gives counter counterexamples that show distinctness for each level illustrated in Figure~\ref{fig: requirements and hierarchy}.

Whenever $U$ is not fully determined by $\vec{Z}, \vec{X}$\footnote{Formally this is described as positivity, i.e. $\Pr(u \given \vec{z}, \vec{x}>0$}, the observed supports of the latent classes overlap. A consequence of this overlap is that identical entries can belong to two different classes. Hence, covariates are insufficient to assign individuals to MTE groups with perfect accuracy. This introduces class-misspecification biases \citep{loh2022evaluating} in addition to invalidating nearest-neighbor approaches \citep{suk2021hybridizing}. To handle this challenge, we develop SPOs as a method of moments.

\subsection{Related Works}\label{sec:related works}
\paragraph{Ladder of Causality} \citet{pearl2009causality} also gives a causal hierarchy: counterfactuals, interventions, associations. We further refine the space in between counterfactuals and interventions: HTEs correspond to counterfactuals that are indexed with respect to a known SCM, and ATEs correspond to interventions.

\paragraph{HTEs} Recent work has studied HTEs for non-overlapping (i.e. qualitatively different) classes \citep{xie2012estimating, wendling2018comparing, kunzel2019causaltoolbox, imai2011estimation}. Popular approaches include causal forests \citep{wager2018estimation} and two-step algorithms generalized by \citet{nie2021quasi}.

\paragraph{MTEs}
MTEs are latent classes defined by heterogeneous treatment effects that overlap in their observed covariates. Early attempts involve learning distinct parameters per latent class \citep{lyu2023estimating, kim2015mixture, kim2015multilevel, suk2021hybridizing}. All of these approaches make use of the unconfoundedness assumption, meaning they do not simultaneously deconfound the treatment effects that they recover (i.e., the latent class may affect outcome, but not treatment). Our approach combines the task of recovering \emph{and} deconfounding heterogeneous treatment effects.

\paragraph{Proximal Causal Inference}
Methods for proximal causal inference deconfound ATEs (but not MTEs) using ``negative controls'' that may depend on the unobserved confounder, but do not influence (at least one of) treatment or outcome \citep{tchetgen2020introduction}.
The majority of work has focused on a continuous unobserved confounder \citep{miao2018confounding,tchetgen2020introduction,mastouri2021proximal}. 

Proximal causal inference and negative controls have demonstrated effective application of the DAG setup from Figure~\ref{fig: requirements and hierarchy} level 4, where $Z$ is a negative control effect and $X$ is a negative control outcome. See \citet{shi2020selective} for a review as well as \citet{negativecontrolcovid} for a recent application to COVID-19 effectiveness. Our method of SPOs can be thought of as a generalization of \citet{miao2018identifying} to joint probabilities instead of conditionals, thereby allowing for continuous covariates. By providing insight into how negative controls synthetically copy potential outcomes, we also give a natural extension to MTEs.

\paragraph{Mixtures}
Latent variable methods have been employed to uncover the joint probability distribution between the latent class $U$, treatment $T$, and outcome $Y$. This approach was popularized by \citep{wang2019blessings}, though criticized by \citet{ogburn2019comment, ogburn2020counterexamples} for not guaranteeing identifiability.

Identifiability of discrete mixture models has been studied extensively. In contrast to clustering approaches, this line of works uses the method of moments to handle overlapping probability densities. The primary focus has been on ``mixtures of products,'' in which the observable variables are conditionally independent from each other when the latent $U$ is held constant. This problem can be thought of as a decomposition into rank 1 tensors, giving rise to a number of tensor-methods spurred by the seminal work of \citet{allman2009identifiability} and later followed up in \citet{anandkumar2014tensor}. Outside of tensor methods, the current best\footnote{in sample and time complexity} algorithm with provable guarantees is developed in \citet{gordon2023identification}. \citet{gordon2023causal} recently generalized this setting to arbitrary DAGs.

\paragraph{Tensor and Matrix Completion}
SPOs are similar in concept to synthetic controls \citep{abadie2010synthetic,abadie2021using}, synthetic interventions \citep{agarwal2020synthetic,squires2022causal}, and synthetic nearest-neighbors \citep{agarwal2023causal}. 
However, these approaches use linear combinations of data \emph{entries}, whereas we use linear combinations of features. We also use higher-order moments and do not attempt to recover any unit-specific information.

\section{PRELIMINARIES}\label{sec:preliminaries}
\paragraph{Random Variables versus Assignments}
We will use the capital Roman alphabet to denote random variables and boldface to denote sets of these random variables. Corresponding lowercase denotes assignment to those random variables. To shorten notation, $X = x$ is sometimes just written as $x$, e.g. $\Pr(y \given x)$.

\paragraph{Vectors of Probabilities and Expectations}
Whenever a quantity is left ambiguous in an expectation or probability, i.e. a boldface $\vec{X}, \vec{Z}$ or a capital $U$, we will first expand it into a column vector and then a matrix if a second ambiguity is given. That is,
\begin{align}
	\vec{E}[\vec{X}] &\coloneq \begin{pmatrix}
	\E(X_1) & \ldots & \E(X_{\abs{\vec{X}}})
	\end{pmatrix}^\top,\\
  \vec{P}[U] &\coloneq \begin{pmatrix}
	\Pr(U = 1) & \ldots &  \Pr(U=k)\\
	\end{pmatrix}^\top,\\
 \vec{E}[X_i \given U] &\coloneq \begin{pmatrix}
	\E(X_i \given U = 1) & \ldots &  \E(X_i \given U=k)\\
	\end{pmatrix}^\top.
\end{align} 

$\vec{E}[\vec{X}Y]$ is a column vector of second-order multilinear moments, while $\vec{E}[ \vec{X}, \vec{Z}] \in \R^{\abs{\vec{X}} \times \abs{\vec{Z}}}$ expands both sets of variables as a matrix:
\begin{equation}
	\vec{E}[\vec{Z}, \vec{X}]_{ij} \coloneq \E(Z_i X_j).
\end{equation}

We will also give vectors of conditional expectations,
\begin{equation}
    \vec{E}[\vec{X} \given t]\coloneq \begin{pmatrix}
	\E(X_1 \given t) & \ldots & \E(X_{\abs{\vec{X}}} \given t)
	\end{pmatrix}^\top.
\end{equation}

$\vec{E}[\vec{X} \given U]$ is a matrix with columns of $\vec{E}[\vec{X} \given u]$, given by
\begin{equation}
	\vec{E}[\vec{X} \given U]_{ij}\coloneq \E(X_i \given U=j).
\end{equation}

\paragraph{Observable Moments} 
Not all of our $\vec{E}[\cdot]$ vectors and matrices can be estimated directly by observable statistics. More specifically, we cannot estimate any vector or matrix of probabilities that involves $U$. To emphasize which matrices are observable, we will replace $\vec{E}$ with $\vec{M}$ for ``observable moments.'' We will similarly replace scalar $\E(\cdot)$ with $M(\cdot)$.

$\vec{M}[\cdot]$ vectors can be decomposed as inner products with $\vec{E}[\cdot \given U]$ vectors and $\vec{P}[U]$.  That is,
\begin{equation} \label{eq: decompose M to P vector}
\vec{M}[\vec{X}] = \vec{E}[\vec{X} \given U] \vec{P}[U].
\end{equation}
When $\vec{X} \indep \vec{Z} \given U$, we have a similar decomposition for second-order moment matrices:
\begin{equation} \label{eq: decompose M to P matrix}
\vec{M}[\vec{X}, \vec{Z}] = \vec{E}[\vec{X} \given U] \diag(\vec{P}[U]) \vec{E}[\vec{Z} \given U]^\top.
\end{equation}

Algorithmic computations will be restricted to using $\vec{M}[\cdot]$s. To prove correctness and analyze stability, we will decompose into $\vec{E}[\cdot]$ and $\vec{P}[\cdot]$, which exist hypothetically but cannot be accessed directly from observed statistics.

\paragraph{Potential Outcomes}
To understand the difference between \emph{intervening} and \emph{conditioning} on $T$, compare the following quantities:
\begin{align}
    \E(Y \given t) &=  \vec{E}[Y \given U, t]^\top \vec{P}[U \given t], \label{eq: condition vec}\\
    \E(Y^{(t)}) &=  \vec{E}[Y \given U, t]^\top \vec{P}[U]. \label{eq: intervention vec}
\end{align}
The crucial difference is that intervening on $T=t$ does change the right column vector. $\E(Y \given t)$ is an observable moment that is estimated by counting $Y=1$ when conditioning on $T=t$. $\E(Y^{(t)})$, in contrast, is not given by an observable moment because we cannot sample from the marginal $U$ while simultaneously drawing from the conditional $Y \given U, t$. This is due to confounding, which shifts the distribution of $U$ between the treatment ($T=1$) and control ($T=0$) groups.

\section{SPOs} \label{sec: SPOs}
For ATEs, our goal is to recover $\E(Y^{(1)} - Y^{(0)})$. We will simplify this using $R \coloneq Y^{(1)} - Y^{(0)}$ for the ``response.'' For MTEs, we must recover $k$ CATEs, i.e. $\vec{E}[R \given U]$, as well as $\vec{P}[U]$. We present SPOs as a method for identifying ATEs and MTEs.
\begin{thm} \label{thm: ATE identifiability}
    The ATE given by $\E(R)$, is identifiable by SPOs if 
    \begin{enumerate}[(i)]\setlength\itemsep{0em}
        \item $\vec{Z} \indep Y \given T, U$
        \item $\vec{X} \indep T \given U$
        \item $\vec{M}[\vec{Z}, \vec{X} \given t]$ is full rank for $t \in {0, 1}$.
    \end{enumerate}
\end{thm}

\begin{thm} \label{thm: MTE identifiability}
    The MTE given by $\vec{E}[R \given U]$ and $\vec{P}[U]$ is identifiable by SPOs if 
    \begin{enumerate}[(i)]\setlength\itemsep{0em}
        \item $\vec{Z} \indep Y \given T, U$
        \item $\vec{X} \indep (Y, T) \given U$
        \item $\vec{M}[\vec{Z}, \vec{X} \given t]$ is full rank for $t \in {0, 1}$.
    \end{enumerate}
\end{thm}
For both theorems, conditions (i) and (ii) correspond to the graphs given in Figure~\ref{fig: requirements and hierarchy}. Notice that $\vec{Z} \indep \vec{X} \given U$ is implied in both cases if no other variables are present (i.e. no other active paths).
\subsection{Main Insight} 
If we could sample from $Y^{(t)}$ \emph{without} conditioning on $T$, we could access the expected value as an observable moment. Notice that if $\vec{X} \indep T \given U$, then $\E[\vec{X} \given U, t] = \E[\vec{X} \given U]$. Our approach will rely on the observation that $\vec{E}[Y^{(t)} \given U]$ is a $k$-dimensional vector that can be expressed as a linear combination of the rows of $\vec{E}[\vec{X} \given U, t]$ provided that the matrix is full-rank, which will be guaranteed by $\vec{M}[\vec{Z}, \vec{X} \given t]$ being full rank due to its decomposition in Equation~\ref{eq: decompose M to P matrix}.
Let this linear combination be given by coefficients $\aalpha$,
\begin{equation} \label{eq: linear combination}
    \vec{E}[Y^{(t)} \given U] = \vec{E}[\vec{X} \given U, t]^\top \aalpha.
\end{equation}
If $\vec{X} \indep T \given U$, the same $\aalpha$ can be applied to the unconditioned moments of $\vec{X}$ to access a ``synthetic potential outcome'':
\begin{equation} \label{eq: get interventional probability}
     \vec{M}[\vec X]^\top \aalpha = \vec{P}[U]^\top \vec{E}[\vec{X} \given U]^\top \aalpha = \E(Y^{(t)}).
\end{equation}
In the next section, we show how to find $\aalpha$.

\subsection{Finding SPOs}
We want to find $\aalpha = (\alpha_1, \ldots, \alpha_k)^\top$ such that, using unconfoundedness,
\begin{equation} \label{eq: linear comb cond on t}
    \vec{E}[Y^{(t)} \given U]^\top = \vec{E}[Y \given U, t]^\top =  \vec{E}[\vec{X} \given U, t]^{\top} \aalpha.
\end{equation}
We begin by recalling that,
\begin{equation}\label{eq: moment decomposition}
    \vec{M}[Y \given t] = \vec{E}[Y \given U, t]^\top \vec{P}[U \given t] = \vec{E}[Y^{(t)} \given U]^\top \vec{P}[U \given t].
\end{equation}
Notice that the left row vector in this inner product is precisely what we want to ``copy''. To compute $\aalpha$, we can set up a system of linear equations that matches second-order moments with $\vec{Z}$:
\begin{equation}\label{eq: solve for a}
    \begin{aligned}
    \vec{M}[\vec{Z}, \vec{X} \given t] \aalpha &= \vec{M}[\vec{Z} Y \given t] \\
    \aalpha &=\vec{M}[\vec{Z}, \vec{X} \given t] ^{-1} \vec{M}[\vec{Z} Y \given t].
\end{aligned}
\end{equation}
To see how this moment-matching approach gives rise to a system of linear equations, apply the expansion given by Equation~\ref{eq: decompose M to P matrix} to $\vec{M}[\vec Z, \vec X]$ on both sides (requires $\vec{Z} \indep \vec{X} \given U)$, and expanding $\vec{M}[\vec{Z}Y \given t]$ similarly using condition (i) from both theorems.
\subsection{ATE Recovery}
The average treatment effect is given by the difference between the moments of the potential outcomes,
\begin{equation}\label{eq: ATE}
    \text{ATE} = \E(R) = E(Y^{(1)}) - E(Y^{(0)}).
\end{equation}
Both moments are accessible using SPO coefficients
\begin{equation}
\begin{aligned}
    \aalpha^{(1)} \coloneq \vec{M}[\vec{Z}, \vec{X} \given T=1]^{-1} \vec{M}[\vec{Z}, Y \given T=1],\\
    \bbeta^{(1)} \coloneq \vec{M}[\vec{Z}, \vec{X} \given T=0] ^{-1} \vec{M}[\vec{Z}, Y \given T=0].
\end{aligned}
\end{equation}
The superscript $1$ is used to signify that the coefficients are used to compute a first order-moment, as we will be generalizing this procedure shortly. We use $\ggamma^{(1)} = \aalpha^{(1)} - \bbeta^{(1)}$ to find the ATE,
\begin{equation}
        \E(R) = \vec{M}[\vec{X}]^\top \aalpha^{(1)} - \vec{M}[\vec{X}]^\top \bbeta^{(1)}  =  \vec{M}[\vec{X}]^\top \ggamma^{(1)}.
\end{equation}

\subsection{MTE Recovery}
We will denote the $\ell$th order moment as 
\begin{equation}
    \E(R^{\odot \ell}) = \vec{P}[U]^\top \vec{E}[R \given U]^{\odot \ell}
\end{equation}
with $\odot \ell$ indicating an element-wise exponent. This quantity is different from $\E(R^{\ell})$, instead being interpreted as the result of choosing $U=u$ and then multiplying $\ell$ i.i.d. samples of $R \given u$.

Information about mixed treatment effects is contained in these higher-order moments of $R$. For example, a homogeneous treatment effect of $0$ will give $\E(R^{\odot 2}) = \E(R) = 0$, whereas a mixed treatment effect of half $+\nicefrac{1}{2}$ and half $-\nicefrac{1}{2}$ will give $\E(R^{\odot 2}) = \nicefrac{1}{4}$, despite an identical $\E(R) = 0$. 

Using $\odot$ to denote element-wise multiplication, we slightly abuse notation to define
\begin{equation}
    \E(Y^{(t)}R^{\odot \ell-1}) \coloneq\left(\vec{E}[Y^{(t)} \given U] \odot \vec{E}[R^{\odot \ell-1} \given U] \right)^\top \vec{P}[U]. \label{eq: elementwise prod}
\end{equation}
We can then expand the $\ell$th order moment as
\begin{equation}
    \E(R^{^{\odot}\ell}) = \E(Y^{(1)}R^{\odot \ell-1}) - \E(Y^{(0)}R^{\odot \ell-1}).
\end{equation}
Hence, our goal is now to estimate $\E(Y^{(t)} R^{\odot \ell - 1})$ the same way we estimated $\E(Y^{(t)})$.
\begin{equation}\label{eq: two copies}
    \begin{aligned}
        \vec{E}[Y^{(1)} \given U] \odot \vec{E}[R^{\odot \ell-1} \given U] &= \vec{E}[\vec{X} \given U, t]^\top \aalpha^{(\ell)},\\
        \vec{E}[Y^{(0)} \given U] \odot \vec{E}[R^{\odot \ell-1} \given U] &= \vec{E}[\vec{X} \given U, t]^\top \bbeta^{(\ell)}.
    \end{aligned}
\end{equation}
 An additional requirement of $Y \indep \vec{X} \given U$ means that
\begin{equation}
    \E[YX_i \given U] = \E[Y \given U] \odot \E[X_i \given U] \; \forall X_i.
\end{equation}
We will compute $\ggamma^{(\ell)}$ when $\ggamma^{(\ell-1)}$ has already been computed (with the base case $\ggamma^{(1)}$ already covered for ATE recovery). Hence, we can assume that $\vec{X}$ can already synthetically ``copy'' $R^{\odot \ell-1}$. We now rewrite Equation~\ref{eq: two copies} using $\ggamma^{(\ell-1)}$ and second-order moments between $X_i$ and $Y$:
\begin{equation}
\begin{aligned}
    \vec{E}[\vec{X} \given U, t]^\top \aalpha^{(\ell)} &=  \vec{E}[Y^{(t)} \given U] \odot \sum_i \ggamma^{(\ell-1)}_i \vec{E}[X_i \given U, t]\\
    &= \sum_i \ggamma^{(\ell-1)}_i\vec{E}[Y X_i \given U, t]
\end{aligned}
\end{equation}
We again moment-match with $\vec{Z}$ to get
\begin{equation}
    \begin{aligned}
        \aalpha^{(\ell)} &= \vec{M}[\vec{Z}, \vec{X} \given T=1]^{-1} \vec{M}[\vec{Z}, \vec{X}Y \given T=1] \ggamma^{(\ell-1)},\\
         \bbeta^{(\ell)} &= \vec{M}[\vec{Z}, \vec{X} \given T=0]^{-1} \vec{M}[\vec{Z}, \vec{X}Y \given T=0] \ggamma^{(\ell-1)}.
    \end{aligned}
\end{equation}
Once we have computed $\ggamma^{(\ell)} = \aalpha^{(\ell)} - \bbeta^{(\ell)}$, we can estimate the $\ell$th moment of the treatment effect,
\begin{equation}
    \E(R^{\odot \ell}) = \vec{M}[\vec{X}]^\top \ggamma^{(\ell)}.
\end{equation}
Using these higher-order moments to recover a multiplicity of treatment effects reduces to the well-studied ``sparse Hausdorff moment problem,'' which seeks to recover $\E[R \given U]$ and $\vec{P}[U]$. Notice that this new mixture model on $R$ only contains the parameters of interest and is therefore no longer beholden to the correct identification of $\Pr(\vec{X}, \vec{Z} \given U)$.
\begin{thm}[\cite{rabani2014learning}]
    $\vec{P}[U]$ and distinct\footnote{i.e. $i \neq j \Rightarrow \E(R \given u_i) \neq \E(R \given u_j)$} $\vec{E}[R \given U]$ can be uniquely identified using $\E(R), \E(R^{\odot 2}), \ldots, \E(R^{\odot 2k-1})$.
\end{thm}
Identifiability of a mixture of $k$ treatment effects requires $2k-1$ moments and can be calculated using Prony's method \citep{de1795essai}, as illustrated and analyzed in \citet{gordon2020sparse}, or the matrix pencil method \citep{hua1990matrix}, which is outlined in Section 2.2 of \citet{kim2019many}. The full algorithm pseudocode is given in Appendix~\ref{sec: alg}. Runtime and sample complexity analysis are detailed in Appendix~\ref{sec: analysis}

\section{THE HIERARCHY OF IDENTIFIABILITY} \label{sec: hierarchy}
We now give an in-depth discussion on the identifiability of each level of our hierarchy and why these levels are distinct. Having just discussed the identifiability of levels 3 and 4, we will briefly explain the identifiability of levels 1 and 2 with slight modifications of previous results. By nature of the hierarchy, a solution for a ``harder'' level can be used to obtain a solution for an ``easier'' level. We will also show the distinctness of each of these levels by giving two different models that resolve as two different ``answers'' for harder levels, but the same ``answer'' for easier levels.

To simplify things, all of our models will use Bernoulli $U, Z, X, T$, and $Y$, with $\vec P[U] = [\nicefrac{1}{2}, \nicefrac{1}{2}]$, and
\begin{equation}
    \vec P[Z = 1 \mid U]
    =
    \vec P[X = 1 \mid U]
    =
    \begin{pmatrix}
        \nicefrac{1}{4}
        \\
        \nicefrac{3}{4}
    \end{pmatrix}.
\end{equation}
The features $\vec{X}$ are therefore $X_1 = \ind(X=0)$ and $X_2 = \ind(X=1)$ and likewise for $\vec{Z}$.
The resulting matrix $\vec M[\vec Z, \vec X]$ is full rank as required by Theorem~\ref{thm: MTE identifiability}.

\paragraph{Level 1: HTEs}
Identifiability of a CATE with respect to an observed adjustment set is perhaps the most well-studied of the levels. In our setup, we have
\begin{equation}
    \E(Y^{(t)} \given u) = \E(Y \given t, u).
\end{equation}
This allows us to compute the CATE as $\E(Y^{(1)} \given u, \vec x, \vec z) - \E(Y^{(0)} \given u, \vec x, \vec z)$, which can be assigned to all data points with $U=u, \vec X = \vec x, \vec Z = \vec z$ when $U$ is observed.

\paragraph{Level 2: Mixtures of Products}
When $U$ is not observed, CATEs cannot be assigned to individual points. For example, if $(\vec{Z}, T, Y, \vec{X})$ are all discrete, then a strictly positive distribution will give nonzero probabilities for $\Pr(\vec z, t, y, \vec{x} \given U=1)$ and $\Pr(\vec z, t, y, \vec{x} \given U=0)$. If the $U=0$ and $U=1$ components exhibit different causal effects, then we can arbitrarily assign all $n$ points with covariates $(\vec z, \vec x)$ to classes $U=0$ and $U=1$ so long as we obey the correct ratio of $\Pr(U=0) : \Pr(U=1)$ in our labeling. Hence, any mixture component overlap makes HTE identifiability at the individual level impossible.

Despite this, under some circumstances, the conditional probability distributions $\Pr(\vec{Z}, T, Y, \vec{X} \given U)$ and $\Pr(U)$ are still identifiable.
One such circumstance is when $\vec Z$, $\vec S$, and $\vec X$ are conditionally independent given $U$, where $\vec S = (T, Y)$; \citep{allman2009identifiability} showed this case to be identifiable, up to symmetries in permuting the labels of $U$, under certain algebraic assumptions on the conditional distribution matrices $\vec P[\vec Z \mid U]$, $\vec P[\vec S \mid U]$, and $\vec P[\vec X \mid U]$.
As we show in Appendix~\ref{apx:moments-levels-2-4}, this result can be extended to include the continuous setting by using moments instead of probabilities.
After recovering the within-component conditional probabilities, $\Pr(\vec{Z}, T, Y, \vec{X} \given U)$, we can also recover $\E(Y^{(t)} \given u) = \E(Y \given t, u)$.

\paragraph{Level 3: MTEs}
SPOs identify MTEs under the conditions given in Theorem~\ref{thm: MTE identifiability}. 
We give two models with invertible moments that identify the same MTE, but represent different mixture models that are not distinguishable at level 2.
In particular, we define two parameterized families:

First, for $\strengthZT \in [0, 1]$, we define
\begin{equation}
    \vec P_\strengthZT[T = 1 \mid Z, U]
    =
    \frac{1}{4}
    \begin{pmatrix}
        3 & 1
        \\
        3 & 1
    \end{pmatrix}
    +
    \frac{\strengthZT}{4}
    \begin{pmatrix}
        0 & 2
        \\
        -2 & 0
    \end{pmatrix}
\end{equation}
Second, for $\strengthXY \in [0, 1]$, we define
\begin{equation}
\begin{aligned}
    \vec P_\strengthXY[Y^{(0)} = 1 \mid X, U]
    &=
    \frac{1}{8}
    \begin{pmatrix}
        7 & 1
        \\
        7 & 1
    \end{pmatrix}
    +
    \frac{\strengthXY}{8}
    \begin{pmatrix}
        0 & 6
        \\
        -6 & 0
    \end{pmatrix}
    \\
    \vec P_\strengthXY[Y^{(1)} = 1 \mid X, U]
    &=
    \frac{1}{8}
    \begin{pmatrix}
        1 & 7
        \\
        1 & 7
    \end{pmatrix}
    +
    \frac{\strengthXY}{8}
    \begin{pmatrix}
        0 & -6
        \\
        6 & 0
    \end{pmatrix}
\end{aligned}
\end{equation}
Thus, for $R = Y^{(1)} - Y^{(0)}$,
\begin{equation}
    \vec E_\strengthXY[R \mid X, U]
    =
    \frac{1}{4}
    \begin{pmatrix}
        -3 & 3
        \\
        -3 & 3
    \end{pmatrix}
    +
    \frac{\strengthXY}{4}
    \begin{pmatrix}
        0 & -6
        \\
        6 & 0
    \end{pmatrix}
\end{equation}
The value of $\strengthZT$ controls how treatment assignment depends on $\vec Z$ and $U$, whereas the value of $\strengthXY$ controls how treatment effect depends on $\vec X$ and $U$.

We define the following two models:
\begin{center}
    \ModelA: \quad$\strengthZT = 0$, $\strengthXY = 0$
    \\
    \ModelB: \quad$\strengthZT = 1$, $\strengthXY = 0$
\end{center}
In both models, since $\strengthXY = 0$, we have $Y \indep X \mid U$, and $\vec E[R \mid U] = [-\nicefrac{3}{4}, \nicefrac{3}{4}]$, i.e., the two models have the same mixture of treatment effects. 
However, for \ModelA, since $\strengthZT = 0$, we have $T \indep Z \mid U$.
Meanwhile, in \ModelB, since $\strengthZT = 1$, we have $T \indep U \mid Z$.
Thus, these two models trade off whether heterogeneity in treatment \emph{assignment} is driven by $U$ or by {\color{blue} $Z \to T$}, and represent two different mixture models.

\begin{figure}
\begin{subfigure}{0.23\textwidth}
    \centering
    \includegraphics[width=\textwidth]{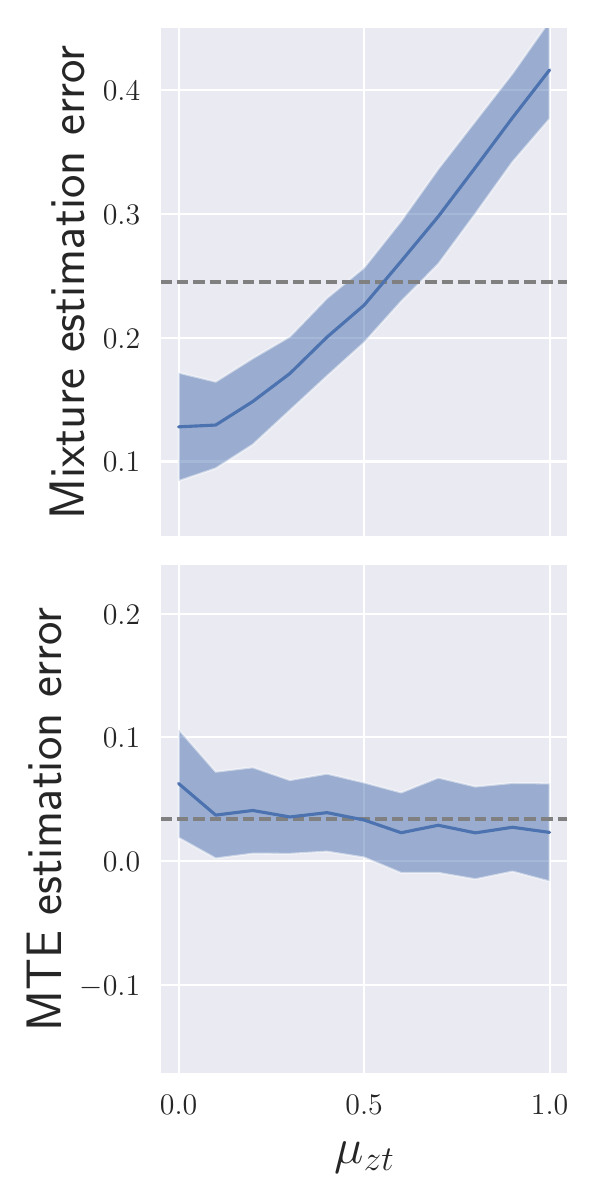}
    \caption{\textbf{Level 2 vs. Level 3}}
    \label{fig:full-mixture-vs-mte}
\end{subfigure}
\begin{subfigure}{0.23\textwidth}
    \centering
    \includegraphics[width=\textwidth]{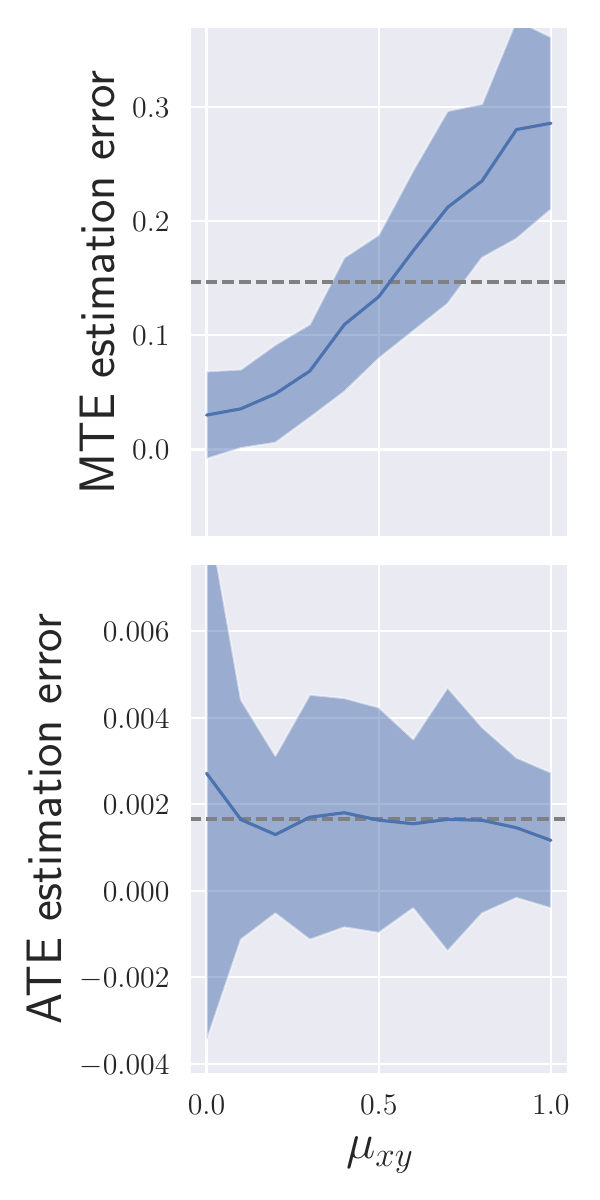}
    \caption{\textbf{Level 3 vs. Level 4}}
    \label{fig:mte-vs-ate}
\end{subfigure}
\caption{
On the left, as we vary $\strengthZT$, mixture estimation error increases but MTE estimation error is stable and close to zero.
On the right, as we vary $\strengthXY$, MTE estimation error increases but MTE estimation error is stable and close to zero.
The blue line is the average error, the shading covers one standard deviation.
The dashed gray line is the mean over all parameter values.
}
\end{figure}

\paragraph{Level 4: ATEs}
SPOs identify ATEs under the conditions given in Theorem~\ref{thm: ATE identifiability}. 
We define a third model:
\begin{center}
    \textbf{\ModelC:}\quad$\strengthZT=1$, $\strengthXY = 1$
\end{center}
In this model, $Y \indep U \mid X$, with $\vec E [R \mid X] = [-\nicefrac{3}{4}, \nicefrac{3}{4}]$.
A quick calculation shows that $\vec E[R \mid U] = [-\nicefrac{3}{8}, \nicefrac{3}{8}]$.
Hence, \ModelC\ has a different mixture of treatment effects than \ModelB, but both models have the same ATE, $\mathbb{E}(R) =  0$.
Thus, these two models trade off whether heterogeneity in treatment \emph{effect} is driven completely by $U$, or completely by {\color{red} $Y \leftarrow X$}.



\section{EMPIRICAL RESULTS}\label{section:empirical-demonstration}

The previous section introduces a model in which the strength of ${\color{blue}Z \leftrightarrow T}$ and ${\color{red} X \rightarrow Y}$ are controlled by $\strengthZT$ and $\strengthXY$. Recall that full mixture recovery (level 2) is identifiable when neither ${\color{blue}Z \leftrightarrow T}$ nor ${\color{red} X \rightarrow Y}$ exist, MTEs (level 3) are identifiable with ${\color{blue}Z \leftrightarrow T}$ but no ${\color{red} X \rightarrow Y}$, and ATEs are identifiable with both ${\color{blue}Z \leftrightarrow T}$ and ${\color{red} X \rightarrow Y}$.

Our empirical results explore the transition between identifiability with respect to $\strengthZT$ and $\strengthXY$, which gradually shift us between the regimes of identifiability for these levels. Each level is computed and evaluated using an algorithm and loss function tailored to the task:
\begin{itemize} \setlength\itemsep{0em}
    \item \textbf{Level 2}: Computed using PARAFAC in \texttt{tensorly} \citep{tensorly} and evaluated using the total variation distance between $\Pr(Z, X, Y, T, U)$ and $\widehat{\Pr}(Z \mid U) \cdot \widehat{\Pr}(X \mid U) \cdot \widehat{\Pr}(Y, T \mid U) \cdot \widehat{\Pr}(U)$.
    \item \textbf{Level 3}: Computed using SPOs (all moments) and the matrix pencil method. Evaluated using the squared $\ell_2$ distance between the true MTE parameters $\vec P[U], \vec E[R \mid U]$ and their corresponding estimates.
    \item \textbf{Level 4}: Computed using SPOs for just the first moment and evaluated by the squared difference between the true ATE $\mathbb{E}(R)$ and its estimate.
\end{itemize}
In all experiments, we report results averaged over 100 runs for each parameter value.
Each single run consists of taking 1,000 samples and running each method method on the empirical moments.
An anonymized repository containing a Python implementation of our SPO method and code for replicating our experiments can be found \href{https://anonymous.4open.science/r/synthetic-potential-outcomes-submission-E900/README.md}{\textcolor{magenta}{here}}.
For mixture recovery, we use the PARAFAC algorithm in \texttt{tensorly} \citep{tensorly} to decompose the empirical probability distribution into factors, then normalize the factors into probability distributions $\widehat{\Pr}(U)$, $\widehat{\Pr}(Z \mid U)$, $\widehat{\Pr}(X \mid U)$ and $\widehat{\Pr}(Y, T \mid U)$.

\paragraph{Level 2 vs. Level 3}
First, we demonstrate the separation between Level 2 and Level 3 by varying the parameter $\strengthZT$ from 0 to 1 while keeping $\strengthXY = 0$, i.e., we interpolate from \ModelA\ to \ModelB.
As shown in Fig.~\ref{fig:full-mixture-vs-mte}, our method of SPOs accurately recovers the mixture of treatment effects for all values of $\strengthZT$.
In contrast, when we perform tensor decomposition, the performance of mixture recovery becomes significantly worse as $\strengthZT$ becomes larger.

\paragraph{Level 3 vs. Level 4}

Next, we demonstrate the separation between Level 3 and Level 4 by varying the parameter $\strengthXY$ from 0 to 1 while keeping $\strengthZT = 1$, i.e., we interpolate from \ModelB\ to \ModelC.
As shown in Fig.~\ref{fig:mte-vs-ate}, our method accurately recovers the average treatment effect for all values of $\strengthXY$; however, as expected, the performance of MTE recovery becomes significantly worse as $\strengthZT$ becomes larger.


\section{DISCUSSION} \label{sec: discussion}
\paragraph{Relaxing Assumptions for Identifiability}
The independence conditions given in Theorem~\ref{thm: ATE identifiability} and Theorem~\ref{thm: MTE identifiability} have a nice graphical correspondence in Figure~\ref{fig: requirements and hierarchy}, but can be relaxed to exchangeability with respect to expectations\footnote{Thank you to the anonymous reviewer who pointed this out to us.}. To understand this, notice that the decomposition given in Equation~\ref{eq: decompose M to P matrix} only demands $\E[XZ]=\E[X \given U]\E[Z \given U]$ for all $Z \in \vec{Z}, X \in \vec{X}$. This is a milder condition than $\vec{X} \indep \vec{Z} \given U$. As a result, Theorem~\ref{thm: ATE identifiability} only requires
\begin{enumerate}[(i)]
    \item $\E[ZY \given U] = \E[Z \given U] \E[Y \given U]$ for all $Z \in \vec{Z}$,
    \item $\E[XT \given U] = \E[X \given U] \E[T \given U]$ for all $X \in \vec{X}$,
\end{enumerate}
for ATE identifiability with SPOs. Similarly, Theorem~\ref{thm: MTE identifiability} only requires
\begin{enumerate}[(i)]
    \item $\E[ZY \given U] = \E[Z \given U] \E[Y \given U]$ for all $Z \in \vec{Z}$,
    \item $\E[XT \given U] = \E[X \given U] \E[T \given U]$ and $\E[XY \given U] = \E[X \given U] \E[Y \given U]$ for all $X \in \vec{X}$,
\end{enumerate}
for MTE identifiability using SPOs.

\paragraph{The Importance of Identifiability} In the absence of validation sets, identifiability is essential. However, it is important to study identifiability at the correct granularity --- too coarse and we limit the information we recover, too fine and we limit the application of the approach. Empirical results show that the identifiability transitions are ``soft,'' meaning that loosely violated requirements can still give accurate results. In Appendix~\ref{sec: analysis}, we give sample complexity results for SPOs relative to the condition number of $\vec{M}[\vec{X}, \vec{Z}]$. Future studies of sample complexity may further resolve stability under mildly violated assumptions.

\paragraph{Mechanism-based Classes} In this paper, we argue that MTEs represent a fundamental granularity worth studying and showed how their identifiability fits within the more well-studied frameworks and problems. MTEs are classes that are differentiated by their mechanisms, which are ensemble-level properties. Covariate values are often insufficient to differentiate relationships between observables, leading to distribution overlap that many clustering approaches cannot appropriately address. It is therefore essential to consider parametric approaches and methods of moments when developing algorithms. These approaches are not only advantageous because of their increased power of identification (as demonstrated by the gap between level 1 and level 2), but also more private, as individuals are never assigned class memberships.

\paragraph{Covariate Richness} Finding a ``sufficiently-rich'' set of features $\vPhi(\vec X), \vTheta(\vec Z)$ is left as an open problem. Many results in proximal causal inference assume that the negative controls are rich categorical variables, e.g. alphabets with cardinality $k$. This can be interpreted with respect to algebraic geometry. While additional parameters usually increase the difficulty of identification, more parameters within $\vec X, \vec Z$ give more degrees of freedom along which these random variables can differ between different classes. Each of these degrees of freedom can be made into a feature, which gives us access to more moments. Because we look at second-order moments between $\vec{Z}, \vec{X}$, a linear increase in parameters corresponds to a quadratic increase in the moments used to recover those parameters. 

\paragraph{Heterogeneity Detection} Our approach is limited by the inherent instability of mixture models. As such, future work should develop statistical tests for the \emph{presence} of (significantly different) MTEs, as in \citet{pearl2022detecting}, rather than guaranteeing full mixture recovery. This problem likely makes up an additional layer of identifiability. Such tests could be used for quality control (e.g. detecting an ineffective batch of medication), quantification of uncertainty in treatment response, and the detection of vaccine-resistant viral variants.


\newpage

\bibliography{allbib}

\begin{thebibliography}{}

\bibitem[Abadie, 2021]{abadie2021using}
Abadie, A. (2021).
\newblock Using synthetic controls: Feasibility, data requirements, and
  methodological aspects.
\newblock {\em Journal of Economic Literature}, 59(2):391--425.

\bibitem[Abadie et~al., 2010]{abadie2010synthetic}
Abadie, A., Diamond, A., and Hainmueller, J. (2010).
\newblock Synthetic control methods for comparative case studies: Estimating
  the effect of california’s tobacco control program.
\newblock {\em Journal of the American statistical Association},
  105(490):493--505.

\bibitem[Agarwal et~al., 2023]{agarwal2023causal}
Agarwal, A., Dahleh, M., Shah, D., and Shen, D. (2023).
\newblock Causal matrix completion.
\newblock In {\em The Thirty Sixth Annual Conference on Learning Theory}, pages
  3821--3826. PMLR.

\bibitem[Agarwal et~al., 2020]{agarwal2020synthetic}
Agarwal, A., Shah, D., and Shen, D. (2020).
\newblock Synthetic interventions.
\newblock {\em arXiv preprint arXiv:2006.07691}.

\bibitem[Allman et~al., 2009]{allman2009identifiability}
Allman, E.~S., Matias, C., and Rhodes, J.~A. (2009).
\newblock Identifiability of parameters in latent structure models with many
  observed variables.

\bibitem[Anandkumar et~al., 2014]{anandkumar2014tensor}
Anandkumar, A., Ge, R., Hsu, D.~J., Kakade, S.~M., Telgarsky, M., et~al.
  (2014).
\newblock Tensor decompositions for learning latent variable models.
\newblock {\em J. Mach. Learn. Res.}, 15(1):2773--2832.

\bibitem[de~Prony, 1795]{de1795essai}
de~Prony, G.~R. (1795).
\newblock Essai experimental et analytique: sur les lois de la dilatabilite des
  fluides elastique et sur celles de la force expansive de la vapeur de l'eau
  et de la vapeur de l'alkool, a differentes temperatures.
\newblock {\em Journal Polytechnique ou Bulletin du Travail fait a l'Ecole
  Centrale des Travaux Publics}.

\bibitem[Gordon et~al., 2021]{gordon2021source}
Gordon, S., Mazaheri, B., Rabani, Y., and Schulman, L. (2021).
\newblock Source identification for mixtures of product distributions.
\newblock In {\em Conference on Learning Theory}, pages 2193--2216. PMLR.

\bibitem[Gordon et~al., 2023]{gordon2023causal}
Gordon, S., Mazaheri, B., Rabani, Y., and Schulman, L. (2023).
\newblock Causal inference despite limited global confounding via mixture
  models.
\newblock In {\em 2nd Conference on Causal Learning and Reasoning}.

\bibitem[Gordon et~al., 2020]{gordon2020sparse}
Gordon, S., Mazaheri, B., Schulman, L.~J., and Rabani, Y. (2020).
\newblock The sparse hausdorff moment problem, with application to topic
  models.
\newblock {\em arXiv preprint arXiv:2007.08101}.

\bibitem[Gordon et~al., 2024]{gordon2023identification}
Gordon, S.~L., Jahn, E., Mazaheri, B., Rabani, Y., and Schulman, L.~J. (2024).
\newblock Identification of mixtures of discrete product distributions in
  near-optimal sample and time complexity.
\newblock In {\em The Thirty Seventh Annual Conference on Learning Theory},
  pages 2071--2091. PMLR.

\bibitem[Hua and Sarkar, 1990]{hua1990matrix}
Hua, Y. and Sarkar, T.~K. (1990).
\newblock Matrix pencil method for estimating parameters of exponentially
  damped/undamped sinusoids in noise.
\newblock {\em IEEE Transactions on Acoustics, Speech, and Signal Processing},
  38(5):814--824.

\bibitem[Imai and Strauss, 2011]{imai2011estimation}
Imai, K. and Strauss, A. (2011).
\newblock Estimation of heterogeneous treatment effects from randomized
  experiments, with application to the optimal planning of the get-out-the-vote
  campaign.
\newblock {\em Political Analysis}, 19(1):1--19.

\bibitem[Imbens and Rubin, 2015]{imbens2015causal}
Imbens, G.~W. and Rubin, D.~B. (2015).
\newblock {\em Causal inference in statistics, social, and biomedical
  sciences}.
\newblock Cambridge university press.

\bibitem[Kendrick Qijun~Li and Tchetgen, 2024]{negativecontrolcovid}
Kendrick Qijun~Li, Xu~Shi, W.~M. and Tchetgen, E.~T. (2024).
\newblock Double negative control inference in test-negative design studies of
  vaccine effectiveness.
\newblock {\em Journal of the American Statistical Association},
  119(547):1859--1870.

\bibitem[Kim and Steiner, 2015]{kim2015multilevel}
Kim, J.-S. and Steiner, P.~M. (2015).
\newblock Multilevel propensity score methods for estimating causal effects: A
  latent class modeling strategy.
\newblock In {\em Quantitative Psychology Research: The 79th Annual Meeting of
  the Psychometric Society, Madison, Wisconsin, 2014}, pages 293--306.
  Springer.

\bibitem[Kim et~al., 2015]{kim2015mixture}
Kim, J.-S., Steiner, P.~M., and Lim, W.-C. (2015).
\newblock Mixture modeling methods for causal inference with multilevel data.
\newblock {\em Advances in multilevel modeling for educational research:
  Addressing practical issues found in real-world applications}, pages
  335--359.

\bibitem[Kim et~al., 2019]{kim2019many}
Kim, Y., Koehler, F., Moitra, A., Mossel, E., and Ramnarayan, G. (2019).
\newblock How many subpopulations is too many? exponential lower bounds for
  inferring population histories.
\newblock In {\em Research in Computational Molecular Biology: 23rd Annual
  International Conference, RECOMB 2019, Washington, DC, USA, May 5-8, 2019,
  Proceedings 23}, pages 136--157. Springer.

\bibitem[Koller and Friedman, 2009]{koller2009probabilistic}
Koller, D. and Friedman, N. (2009).
\newblock {\em Probabilistic graphical models: principles and techniques}.
\newblock MIT press.

\bibitem[Kossaifi et~al., 2019]{tensorly}
Kossaifi, J., Panagakis, Y., Anandkumar, A., and Pantic, M. (2019).
\newblock Tensorly: Tensor learning in python.
\newblock {\em Journal of Machine Learning Research}, 20(26):1--6.

\bibitem[Kruskal, 1977]{kruskal1977three}
Kruskal, J.~B. (1977).
\newblock Three-way arrays: rank and uniqueness of trilinear decompositions,
  with application to arithmetic complexity and statistics.
\newblock {\em Linear algebra and its applications}, 18(2):95--138.

\bibitem[K{\"u}nzel et~al., 2019]{kunzel2019causaltoolbox}
K{\"u}nzel, S.~R., Walter, S.~J., and Sekhon, J.~S. (2019).
\newblock Causaltoolbox—estimator stability for heterogeneous treatment
  effects.
\newblock {\em Observational Studies}, 5(2):105--117.

\bibitem[Lauritzen, 1996]{lauritzen1996graphical}
Lauritzen, S.~L. (1996).
\newblock {\em Graphical models}, volume~17.
\newblock Clarendon Press.

\bibitem[Loh and Kim, 2022]{loh2022evaluating}
Loh, W.~W. and Kim, J.-S. (2022).
\newblock Evaluating sensitivity to classification uncertainty in latent
  subgroup effect analyses.
\newblock {\em BMC Medical Research Methodology}, 22(1):247.

\bibitem[Lyu et~al., 2023]{lyu2023estimating}
Lyu, W., Kim, J.-S., and Suk, Y. (2023).
\newblock Estimating heterogeneous treatment effects within latent class
  multilevel models: A bayesian approach.
\newblock {\em Journal of Educational and Behavioral Statistics}, 48(1):3--36.

\bibitem[Mastouri et~al., 2021]{mastouri2021proximal}
Mastouri, A., Zhu, Y., Gultchin, L., Korba, A., Silva, R., Kusner, M., Gretton,
  A., and Muandet, K. (2021).
\newblock Proximal causal learning with kernels: Two-stage estimation and
  moment restriction.
\newblock In {\em International conference on machine learning}, pages
  7512--7523. PMLR.

\bibitem[Miao et~al., 2018]{miao2018identifying}
Miao, W., Geng, Z., and Tchetgen~Tchetgen, E.~J. (2018).
\newblock Identifying causal effects with proxy variables of an unmeasured
  confounder.
\newblock {\em Biometrika}, 105(4):987--993.

\bibitem[Miao et~al., 2024]{miao2018confounding}
Miao, W., Shi, X., Li, Y., and Tchetgen~Tchetgen, E.~J. (2024).
\newblock A confounding bridge approach for double negative control inference
  on causal effects.
\newblock {\em Statistical Theory and Related Fields}, pages 1--12.

\bibitem[Nie and Wager, 2021]{nie2021quasi}
Nie, X. and Wager, S. (2021).
\newblock Quasi-oracle estimation of heterogeneous treatment effects.
\newblock {\em Biometrika}, 108(2):299--319.

\bibitem[Ogburn et~al., 2019]{ogburn2019comment}
Ogburn, E.~L., Shpitser, I., and Tchetgen, E. J.~T. (2019).
\newblock Comment on “blessings of multiple causes”.
\newblock {\em Journal of the American Statistical Association},
  114(528):1611--1615.

\bibitem[Ogburn et~al., 2020]{ogburn2020counterexamples}
Ogburn, E.~L., Shpitser, I., and Tchetgen, E. J.~T. (2020).
\newblock Counterexamples to" the blessings of multiple causes" by wang and
  blei.
\newblock {\em arXiv preprint arXiv:2001.06555}.

\bibitem[Pearl, 2009]{pearl2009causality}
Pearl, J. (2009).
\newblock {\em Causality}.
\newblock Cambridge university press.

\bibitem[Pearl, 2022]{pearl2022detecting}
Pearl, J. (2022).
\newblock Detecting latent heterogeneity.
\newblock In {\em Probabilistic and causal inference: The works of judea
  pearl}, pages 483--506.

\bibitem[Peters et~al., 2017]{peters2017elements}
Peters, J., Janzing, D., and Sch{\"o}lkopf, B. (2017).
\newblock {\em Elements of causal inference: foundations and learning
  algorithms}.
\newblock The MIT Press.

\bibitem[Rabani et~al., 2014]{rabani2014learning}
Rabani, Y., Schulman, L.~J., and Swamy, C. (2014).
\newblock Learning mixtures of arbitrary distributions over large discrete
  domains.
\newblock In {\em Proceedings of the 5th conference on Innovations in
  theoretical computer science}, pages 207--224.

\bibitem[Rosenbaum and Rubin, 1983]{rosenbaum1983central}
Rosenbaum, P.~R. and Rubin, D.~B. (1983).
\newblock The central role of the propensity score in observational studies for
  causal effects.
\newblock {\em Biometrika}, 70(1):41--55.

\bibitem[Shi et~al., 2020]{shi2020selective}
Shi, X., Miao, W., and Tchetgen, E.~T. (2020).
\newblock A selective review of negative control methods in epidemiology.
\newblock {\em Current epidemiology reports}, 7:190--202.

\bibitem[Simpson, 1951]{simpson1951interpretation}
Simpson, E.~H. (1951).
\newblock The interpretation of interaction in contingency tables.
\newblock {\em Journal of the Royal Statistical Society: Series B
  (Methodological)}, 13(2):238--241.

\bibitem[Squires et~al., 2022]{squires2022causal}
Squires, C., Shen, D., Agarwal, A., Shah, D., and Uhler, C. (2022).
\newblock Causal imputation via synthetic interventions.
\newblock In {\em Conference on Causal Learning and Reasoning}, pages 688--711.
  PMLR.

\bibitem[Suk et~al., 2021]{suk2021hybridizing}
Suk, Y., Kim, J.-S., and Kang, H. (2021).
\newblock Hybridizing machine learning methods and finite mixture models for
  estimating heterogeneous treatment effects in latent classes.
\newblock {\em Journal of Educational and Behavioral Statistics},
  46(3):323--347.

\bibitem[Tchetgen et~al., 2020]{tchetgen2020introduction}
Tchetgen, E. J.~T., Ying, A., Cui, Y., Shi, X., and Miao, W. (2020).
\newblock An introduction to proximal causal learning.
\newblock {\em arXiv preprint arXiv:2009.10982}.

\bibitem[Wager and Athey, 2018]{wager2018estimation}
Wager, S. and Athey, S. (2018).
\newblock Estimation and inference of heterogeneous treatment effects using
  random forests.
\newblock {\em Journal of the American Statistical Association},
  113(523):1228--1242.

\bibitem[Wang and Blei, 2019]{wang2019blessings}
Wang, Y. and Blei, D.~M. (2019).
\newblock The blessings of multiple causes.
\newblock {\em Journal of the American Statistical Association},
  114(528):1574--1596.

\bibitem[Wendling et~al., 2018]{wendling2018comparing}
Wendling, T., Jung, K., Callahan, A., Schuler, A., Shah, N.~H., and Gallego, B.
  (2018).
\newblock Comparing methods for estimation of heterogeneous treatment effects
  using observational data from health care databases.
\newblock {\em Statistics in medicine}, 37(23):3309--3324.

\bibitem[Xie et~al., 2012]{xie2012estimating}
Xie, Y., Brand, J.~E., and Jann, B. (2012).
\newblock Estimating heterogeneous treatment effects with observational data.
\newblock {\em Sociological methodology}, 42(1):314--347.

\end{thebibliography}

\section*{Checklist}

 \begin{enumerate}

 \item For all models and algorithms presented, check if you include:
 \begin{enumerate}
   \item A clear description of the mathematical setting, assumptions, algorithm, and/or model. [Yes]
   \item An analysis of the properties and complexity (time, space, sample size) of any algorithm. [Yes]
   \item (Optional) Anonymized source code, with specification of all dependencies, including external libraries. [Yes]
 \end{enumerate}

 \item For any theoretical claim, check if you include:
 \begin{enumerate}
   \item Statements of the full set of assumptions of all theoretical results. [Yes]
   \item Complete proofs of all theoretical results. [Yes]
   \item Clear explanations of any assumptions. [Yes]     
 \end{enumerate}

 \item For all figures and tables that present empirical results, check if you include:
 \begin{enumerate}
   \item The code, data, and instructions needed to reproduce the main experimental results (either in the supplemental material or as a URL). [Yes/No/Not Applicable]
   \item All the training details (e.g., data splits, hyperparameters, how they were chosen). [Yes]
         \item A clear definition of the specific measure or statistics and error bars (e.g., with respect to the random seed after running experiments multiple times). [Yes]
         \item A description of the computing infrastructure used. (e.g., type of GPUs, internal cluster, or cloud provider). [Yes]
 \end{enumerate}

 \item If you are using existing assets (e.g., code, data, models) or curating/releasing new assets, check if you include:
 \begin{enumerate}
   \item Citations of the creator If your work uses existing assets. [Not Applicable]
   \item The license information of the assets, if applicable. [Not Applicable]
   \item New assets either in the supplemental material or as a URL, if applicable. [Yes]
   \item Information about consent from data providers/curators. [Not Applicable]
   \item Discussion of sensible content if applicable, e.g., personally identifiable information or offensive content. [Not Applicable]
 \end{enumerate}

 \item If you used crowdsourcing or conducted research with human subjects, check if you include:
 \begin{enumerate}
   \item The full text of instructions given to participants and screenshots. [Not Applicable]
   \item Descriptions of potential participant risks, with links to Institutional Review Board (IRB) approvals if applicable. [Not Applicable]
   \item The estimated hourly wage paid to participants and the total amount spent on participant compensation. [Not Applicable]
 \end{enumerate}

 \end{enumerate}

\newpage
\onecolumn
\appendix

\section{Algorithm Pseudocode} \label{sec: alg}
\begin{algorithm}
\DontPrintSemicolon
\SetKwFunction{NextMoment}{NextMomentCoeffs}
\SetKwFunction{FirstMoment}{FirstMomentCoeffs}
\SetKwFunction{PronyOrPencil}{PronyOrPencil}
\SetKwProg{Fn}{Function}{:}{}

\KwIn{$\vec{M}[\cdot]$ matrices computed by estimating conditional expected values.}
\KwResult{$\ggamma^{(1)}, \ldots, \ggamma^{(2k-1)}$}

\tcc{Compute $\ggamma^{(1)}$ coefficients for $\vec{X}$ using reference $\vec{Z}$.}

\Fn{\FirstMoment{$\vec{M}[\vec{Z}, \vec{X} \given T]$, $\vec{M}[\vec{Z}, Y \given T]$}}{
        $\aalpha^{(1)} \gets \vec{M}[\vec{Z}, \vec{X} \given T=1]^{-1} \vec{M}[\vec{Z}, Y \given T=1]$\;
        $\bbeta^{(1)} \gets \vec{M}[\vec{Z}, \vec{X} \given T=0] ^{-1} \vec{M}[\vec{Z}, Y \given T=0]$\;
        \KwRet $\aalpha^{(1)} - \bbeta^{(1)}$\;
}

\tcc{Compute $\ggamma^{(\ell)}$ for $\vec{X}$ using reference $\vec{Z}$ and previous $\ggamma^{(\ell - 1)}$.}
\Fn{\NextMoment{$\vec{M}[\vec{Z}, \vec{X} \given T]$, $\vec{M}[\vec{Z}, \vec{X}Y \given T]$, $\ggamma^{(\ell-1)}$}}{
        $\aalpha^{(\ell)} \gets \vec{M}[\vec{Z}, \vec{X} \given T=1]^{-1} \vec{M}[\vec{Z}, \vec{X}Y \given T=1] \ggamma^{(\ell-1)}$\;
        $\bbeta^{(\ell)} \gets \vec{M}[\vec{Z}, \vec{X} \given T=0]^{-1} \vec{M}[\vec{Z}, \vec{X}Y \given T=0] \ggamma^{(\ell-1)}$\;
        \KwRet $\aalpha^{(\ell)} - \bbeta^{(\ell)}$\;
}

$\ggamma^{(1)} \gets $ \FirstMoment{$\vec{M}[\vec{Z}, \vec{X} \given T]$, $\vec{M}[\vec{Z}, Y \given T]$}\;
$\nu_1 \gets \vec{M}[\vec{X}]^{\top}\gamma^{(1)}$\;
\For{$\ell \gets2$ \KwTo $2k-1$}{
  $\ggamma^{(\ell)} \gets $ \NextMoment{$\vec{M}[\vec{Z}, \vec{X} \given T]$, $\vec{M}[\vec{Z}, \vec{X}Y \given T]$, $\ggamma^{(\ell-1)}$}\;
  $\nu_\ell \gets \vec{M}[\vec{X}]^{\top}\gamma^{(\ell)}$\;
}

$\vec{P}[U], \vec{E}[R \given U] \gets$ \PronyOrPencil{$\nu_1, \ldots, \nu_{2k-1}$}

\caption{Recovers MTEs using SPOs.}\label{alg: SPOs}
\end{algorithm}

\section{Analysis}\label{sec: analysis}
In this section we will analyze the sample and time complexity of SPOs. The time complexity of computing an SPO is $\mathcal{O}(k^4)$, which is mild compared to the time required to process the approach's data-demands. For this reason, the \emph{sample} complexity is the dominating factor when it comes to run-time.
\subsection{Time Complexity}
The computation of an SPO involves inverting a $k \times k$ matrix and multiplying it. This can be done in $\mathcal{O}(k^3)$ time. When computing higher order moments of the treatment effect, we will pick up another factor or $k$. The final step of applying Matrix pencil or Prony's method involves solving for an eigensystem, which is again $\mathcal{O}(k^3)$. All together, the time complexity becomes $\mathcal{O}(k^4)$, which is relatively mild when considering the time involved in processing the data.

\subsection{Sample Complexity}
The sample complexity for a single SPO computation is given in Theorem~\ref{thm: sample complexity of one synthetic bit}.

\begin{thm}[SPO Sample Complexity] \label{thm: sample complexity of one synthetic bit}
    Let $\pi := \min_t \Pr(T=t)$ and let $\lambda_k$ be the smallest eigenvalue of $\vec{E}[\vec{X} \given U]$. The sample complexity of calculating $\E(Y^{(t)})$ using SPOs is $\mathcal{O}(k^6 \pi^{-1} \lambda_{k}^{-2})$.
\end{thm}
The formal proof is delayed to Appendix~\ref{apx: sample complexity}. We will now discuss a few notable aspects of the sample complexity.

First, the dependence on $k^6$ shows the sample-complexity-dependent nature of this approach -- the time needed to compute a result will primarily scale with the data needed to accurately compute a solution. The sample complexity's dependence on $\pi^{-1}$ shows how computing SPOs for rare treatments scales in difficulty, as it limits the data that we have to be able to calculate $\aalpha$'s to compute the SPO. 

The primary veil of sample complexity lies in $\lambda_k^{-2}$, which essentially says that $\vec{E}[\vec{X} \given U]$ must be well-conditioned. Fortunately, the condition number of $\vec{M}[\vec{Z}, \vec{X} \given t]$ can be checked in practice to ensure a well-conditioned $\vec{E}[\vec{X} \given U]$.

\paragraph{Higher Order Moments}
To recover MTEs, we must use solutions for SPOs at lower orders to bootstrap higher-order moments. This process accumulates errors, particularly with respect to the dependence on the condition number of $\vec{E}[\vec{X} \given U]$. \citet{gordon2021source} outlines an approach that involves only $\mathcal{O}(\log(k))$ steps within this chain, which involves incrementing $\ell$ by more than $1$ using \emph{two} synthetic copies (requiring another $\vec{X}'$ with $\vec{X}' \indep (Y, T) \given U$).

\citet{gordon2020sparse} and \cite{kim2019many} outline the sample complexities of Prony's and the matrix pencil method respectively. A notable factor is a minimum difference between treatment effects across components --- clearly sub-populations with identical or close-to-identical treatment effects will be hard to disentangle. The stability is similarly dependent on the lowest probability sub-population, as rare populations require more data to resolve. We note that these dependencies are not of serious concern to practitioners who are primarily concerned with the \emph{existence} of nontrivial differences in MTEs as a motivator for further investigation.

\section{Sample Complexity Proof} \label{apx: sample complexity}
We will now provide an analysis of the empirical error propagation for our method. $\vec{E}$ vectors and matrices can be estimated empirically on a dataset $D$, denoted $\hat{\vec{E}}^{D}[\cdot]$. Empirical estimates of all $\hat{M}[\cdot]$ can be directly estimated. We will also use $\hat{P}[\cdot], \vec{E}[\cdot]$ vectors and matrices that cannot be directly observed, but still exist hypothetically for anyone who observes $U$.

Equation~\ref{eq: solve for a} tells us how to compute the coefficients for a synthetic bit in the presence of perfect statistics. In reality, the computation we will be performing is on imperfect empirical estimates,
\begin{equation} \label{eq: solve for a empirically}
    \hat{\boldsymbol{\alpha}} = \hat{\vec{M}}[\vec{Z}, \vec{X} \given t] ^{-1} \hat{\vec{M}}[\vec{Z}Y \given t].
\end{equation}
Expanding these moments into unobservable $\tvec{P}[\cdot]$ matrices using Equation~\ref{eq: decompose M to P matrix} gives
\begin{equation}
\begin{aligned}
    \tvec{M}[\vec{Z}, \vec{X} \given t] &= \tvec{E}[\vec{Z} \given U, t] \diag(\tvec{P}[U \given t]) \tvec{E}[\vec{X} \given U, t]^\top,\\
    \tvec{M}[\vec{Z}, Y \given t] &= \tvec{E}[\vec{Z} \given U, t] \diag(\tvec{P}[U \given t]) \tvec{E}[Y \given U, t]^\top.\\
\end{aligned}
\end{equation}
Substituting these expansions and recalling that $\vec{X} \indep T \given U$ simplifies Equation~\ref{eq: solve for a empirically},
\begin{equation} \label{eq: solve for a empirically simplified}
    \hat{\boldsymbol{\alpha}}^\top = \tvec{E}[Y \given U, t]^\top \tvec{E}[\vec{X} \given U, t]^{-1} = \tvec{E}[Y \given U, t]^\top \tvec{E}[\vec{X} \given U]^{-1}.
\end{equation}

An important observation from Equation~\ref{eq: solve for a empirically simplified} is that $\hat{\boldsymbol{\alpha}}$ no longer depends on $\vec{Z}$. That is, we do not have to worry about the stability of sampled statistics from $\vec{Z}$ because any noise is canceled out by the inversion (so long as we do not deviate into non-invertable $\tvec{P}[\vec{Z} \given U, t]$). This significantly simplifies our analysis.

We will analyze the stability of potential outcome expectation estimations computed using the empirical version of Equation~\ref{eq: get interventional probability}. We will distinguish between the original dataset $D$ and the post-selected $D_t$, which is formed from the subset of points for which $T=t$. We therefore want to analyze
\begin{equation} \label{eq: solve for do empirically}
    \delta \hat{\E}(Y^{(t)}) = \tvec{M}^{D}[\vec{X}]^\top \hat{\boldsymbol{\alpha}}^{D_t},
\end{equation}
where we use $\delta$ to denote the difference between the true and empirical values, i.e. $\ddelta\tvec{E}^{D}[\cdot] := \tvec{E}^{D}[\cdot] - \vec{E}[\cdot]$. Statistics computed on the $D_t$ dataset will be less accurate than those on the full $D$ dataset because of the smaller sample-size, so we have expressed our sample complexity in terms of $\min_i \vec{P}[T]_i$.

\subsection{Proof of Theorem~\ref{thm: sample complexity of one synthetic bit}}
To analyze the sample complexity of synthetic potential outcomes, we will give a probably approximately correct bound by separately (1) analyzing the propagation of errors from observed statistics to the recovered $\E[Y^{(t)}]$ and (2) determining the data needed to keep those errors in check. To tackle (1), we will work under Assumption~\ref{assume: eps}.
\begin{assum}\label{assume: eps}
    For all $V \in \vec{X} \cup \vec{Z} \cup \{T, Y\}$ we assume $\delta \hat{\E}^{D}(V), \delta \hat{\E}^{D_t}(V) \leq \varepsilon$. Similarly we assume $\delta \hat{\Pr}^{D}(U), \delta \hat{\Pr}^{D_t}(U) \leq \varepsilon$.
\end{assum}
Lemma~\ref{lem: error in terms of norm E} gives us an upper bound on the error in a synthetic potential outcome.
\begin{lem} \label{lem: error in terms of norm E}
Under Assumption~\ref{assume: eps},
\begin{equation}
\delta\hat{\E}(Y^{(t)}) \leq 2 k^4 \varepsilon \norm{\tvec{E}^{D_t}[\vec{X} \given U]^{-1}}  + 2 k^2 \varepsilon + k^2 \varepsilon^2.
\end{equation}
\end{lem}

\begin{proof}
First expand Equation~\ref{eq: solve for do empirically},
  \begin{equation} \label{eq: expansion for epeirical estimate}
      \hat{\E}(Y^{(t)}) = \tvec{E}^{D_t}[Y \given U, t]^\top \tvec{E}^{D_t}[\vec{X} \given U]^{-1} \tvec{E}^{D}[\vec{X} \given U] \tvec{P}^D[U].
  \end{equation}
Observe that $\tvec{E}^{D_t}[\vec{X} \given U]^{-1} \tvec{E}^{D}[\vec{X} \given U]$ involves \emph{two different empirical estimates} of $\vec{E}[\vec{X} \given U]$, one of which is inverted. Define the ``error matrix,''
\begin{equation}
    \vec{\Delta} \coloneq \tvec{E}^{D}[\vec{X} \given U] - \tvec{E}^{D_t}[\vec{X} \given U],
\end{equation}
and expand to get
\begin{equation}
      \tvec{E}^{D_t}[\vec{X} \given U]^{-1} \tvec{E}^{D}[\vec{X} \given U] = \tvec{E}^{D_t}[\vec{X} \given U]^{-1}(\tvec{E}^{D_t}[\vec{X} \given U] + \vec{\Delta}) = \vec{I} + \tvec{E}^{D_t}[\vec{X} \given U]^{-1} \vec{\Delta}.
\end{equation}
Recall that $\E(Y^{(t)}) = \delta \vec{E}[Y \given U, t]^\top \delta \vec{P}[U]$ and write the empirical computation in terms of the ``correct'' (in {\color{blue} blue}) values plus their errors (in {\color{red} red}),
\begin{align}\label{eq: correct and errors}
      \hat{\E}(Y^{(t)}) &= ({\color{blue} \vec{E}[Y \given U, t]^\top} + {\color{red} \delta \tvec{E}^{D_t}[Y \given U, t]^\top}) ({\color{blue} \vec{I}} + {\color{red} \tvec{E}^{D_t}[\vec{X} \given U]^{-1} \vec{\Delta}}) ({\color{blue}\vec{P}[U]} + {\color{red}\delta \tvec{P}^D[U]}),
\end{align}
This gives us an expression for the difference between the true and computed do-intervention,
\begin{equation} \label{eq: error expanded}
    \begin{aligned}
    \delta\hat{\E}(Y^{(t)}) =& \tvec{E}^{D_t}[Y \given U, t]^\top \tvec{E}^{D_t}[\vec{X} \given U]^{-1} \vec{\Delta} \tvec{P}^D[U] + \vec{E}[Y \given U, t]^\top \delta \tvec{P}^D[U] \\
    &+ \delta \tvec{E}^{D_t}[Y \given U, t]^\top \vec{P}[U] + \delta \tvec{E}^{D_t}[Y \given U, t]^\top \delta \tvec{P}^D[U].
\end{aligned}    
\end{equation}
In order to upper bound the error in Equation ~\ref{eq: error expanded} using Cauchy-Schwartz, we use Assumption~\ref{assume: eps} to find the following bounds on norms.
\begin{align}
    \norm{\tvec{E}^{D_t}[Y \given U, t]} &\leq k\\
    \norm{\vec \Delta} &\leq 2 k^2 \varepsilon\\
    \norm{\tvec{P}[U]} &\leq k\\
    \norm{\vec{E}^{D_t}[Y \given U, t]} &\leq k\\
    \norm{\delta \tvec{P}^{D}[U]} &\leq k \varepsilon\\
    \norm{\delta \tvec{E}^{D_t}[Y \given U, t]} &\leq k \varepsilon\\
    \norm{\vec{P}[U]} &\leq k
\end{align}
Applying these inequalities gives
\begin{equation} \label{eq: 48 end bound}
    \delta\hat{\E}(Y^{(t)}) \leq 2 k^3 \varepsilon \norm{\tvec{E}^{D_t}[\vec{X} \given U]^{-1}}  + 2 k^2 \varepsilon + k^2 \varepsilon^2. \qedhere
\end{equation}
\end{proof}
All that remains is to upper bound $\norm{\tvec{P}^{D_t}[\vec{X} \given U]^{-1}}$. Lemma~\ref{lem: bound norm} shows how this is related to the condition number of $\vec{E}[\vec{X} \given U]$.
\begin{lem} \label{lem: bound norm}
    Let $\lambda_k$ be the smallest eigenvalue of $\vec{E}^{D_t}[\vec{X} \given U]$. Then,
    \begin{equation}
        \norm{\tvec{E}^{D_t}[\vec{X} \given U]^{-1}} \leq \frac{1}{\lambda_k - k^2 \varepsilon}.
    \end{equation}
\end{lem}
\begin{proof}
The operator norm of an the inverse of $\tvec{E}^{D_t}[\vec{X} \given U]$ can be upper bounded by $1/\lambda_k$ where $\hat{\lambda_k}$ is the smallest eigenvalue of $\tvec{E}^{D_t}[\vec{X} \given U]$. We know that $\lambda_k - \hat{\lambda}_k$ cannot be any greater than the operator norm of the difference, $\norm{\delta \tvec{E}^{D_t}(\vec{X} \given U)} \leq k \varepsilon$. Therefore we have
\begin{equation}
    \norm{\tvec{E}^{D_t}[\vec{X} \given U]^{-1}} \leq \frac{1}{\lambda_k - k \varepsilon}. \qedhere
\end{equation}
\end{proof}

The proof of Theorem~\ref{thm: sample complexity of one synthetic bit} now involves analyzing the sample complexity needed to achieve Assumption~\ref{assume: eps} with an $\varepsilon$ such that $\delta \hat{\Pr}(Y \given \Do(t))$ does not depend on $k$ or $\lambda_k$. Hoeffding's inequality gives an upper bound on the probability of failing Assumption~\ref{assume: eps} for an arbitrary variable $V$ on $n$ samples:
\begin{equation}
    \Pr(\abs{\hat{V} - \E[V]} \geq \varepsilon) \leq 2 \exp(-2 \varepsilon^2 n).
\end{equation}
In order to ensure that this upper bound remains constant with respect to $k$ and $\lambda_k$ we must have $n$ proportional to $1/\varepsilon^2$. For our error $\delta \hat{\Pr}(Y \given \Do(t))$ to remain constant, Lemmas ~\ref{lem: error in terms of norm E} and ~\ref{lem: bound norm} require $\varepsilon \leq \frac{\lambda_k - k^2}{k^4}$, which consequently gives the sample complexity in Theorem~\ref{thm: sample complexity of one synthetic bit} (other than the $\pi^{-1}$, which is due to the data-constraints of conditioning on treatment, as explained in the main paper).

\section{Mixture and ATE identifiability with moments}\label{apx:moments-levels-2-4}

\newcommand{\vw}{\vec{w}}
\newcommand{\vB}{\vec{B}}
\newcommand{\vC}{\vec{C}}
\newcommand{\vE}{\vec{E}}
\newcommand{\vM}{\vec{M}}
\newcommand{\vT}{\vec{T}}
\newcommand{\vP}{\vec{P}}
\newcommand{\krank}{\textnormal{rank}_K}
\newcommand{\tvY}{\tilde{\vec{Y}}}

\subsection{Mixture identifiability with moments (Level 2)}

Here, we generalize the proof of mixture identifiability from \cite{allman2009identifiability} to also include continuous-valued feature maps.
We recall that the \textit{Kruskal rank} of a matrix $M$, denoted $\krank(M)$, is the largest number $r$ such that all sets of $r$ columns in $M$ are linearly independent.

We use the following notation for tensors: given $\vw \in \mathbb{R}^k$, $\vA \in \mathbb{R}^{a \times k}$, $\vB \in \mathbb{R}^{b \times k}$, and $\vC \in \mathbb{R}^{c \times k}$, we define the order-3 tensor $\vT = [\vw; \vA, \vB, \vC] \in \mathbb{R}^{a \times b \times c}$ as
\[
\vT_{ij\ell} = \sum_{k'=1}^k w_{k'} \cdot A_{ik'} \cdot B_{jk'} \cdot C_{\ell k'}
\]

Recall from Section~\ref{sec:preliminaries} the definitions for the matrices $\vE[\vec Z \mid U] \in \mathbb{R}^{d_1 \times k}$ and $\vE[\vec X \mid U] \in \mathbb{R}^{d_2 \times k}$:
\[
\vE[\vec Z \mid U]_{iu} = \E (Z_i \mid U = u)
\]
\[
\vE[\vec X \mid U]_{ju} = \E (X_j \mid U = u)
\]
Define a new random vector $\vec S = (\kron_{T = 0}, \kron_{T = 1}, Y \kron_{T = 0}, Y \kron_{T = 1})$, and a new matrix $\vE[\vec S \mid U] \in \mathbb{R}^{4 \times k}$:
\[
\vE[\vec S \mid U]_{\ell u} = \E (S_\ell \mid U = u)
\]

We define the third-order tensor of moments $\vM[\vZ, \vX, \vec S] \in \mathbb{R}^{d_1 \times d_2 \times 4}$
\[
    \vM[\vec Z, \vec X, \vec S]_{ijt} 
    = 
    \E[Z_i \cdot X_j \cdot S_\ell]
\]

\begin{lem}
    Let $\vec Z$, $\vec X$, and $\vec S$ be independent given $U$.
    Let
    \[
    I_1 = \krank(\vE[\vec Z \mid U]),
    \quad
    I_2 = \krank(\vE[\vec X \mid U]),
    \quad
    \textrm{and}
    \quad
    I_3 = \krank(\vE[\vec S \mid U]).
    \]
    Assume that $I_1 + I_2 + I_3 \geq 2k + 2$.
    Then, given $\vM[\vZ, \vX, \vec S]$, we can generically identify $\vP(U)$, $\vE[\vec Z \mid U]$, $\vE[\vec X \mid U]$, and $\vE[\vec S \mid U]$ up to permutation of the labels of $U$.
\end{lem}
\begin{proof}
    For convenience, we define the following extensions:
    \begin{align*}
        \vZ' &= [Z_1, Z_2, \ldots, Z_{d_1}, 1],
        \\
        \vX' &= [X_1, X_2, \ldots, X_{d_2}, 1], \textrm{and}
        \\
        \vec S' &= [S_1, S_2, S_3, S_4, 1],
    \end{align*}
    along with
    \begin{align*}
        \vE[\vec Z' \mid U]_{iu} &= \E (Z'_i \mid U = u),
        \\
        \vE[\vec X' \mid U]_{ju} &= \E (X'_j \mid U = u), \textrm{and}
        \\
        \vE[\vec S' \mid U]_{\ell u} &= \E (S'_\ell \mid U = u),
    \end{align*}
    and finally, $\vM[\vec Z', \vec X', \vec S'] \in \mathbb{R}^{(d_1 + 1) \times (d_2 + 1) \times 5}$ as
    \[
    \vM[\vec Z', \vec X', \vec S']_{ij\ell}
    =
    \E[Z'_i \cdot X'_j \cdot S'_\ell].
    \]

    Note that $\vE[\vec Z' \mid U]$ is equal to $\vE[\vec Z \mid U]$ concatenated with an additional row of $1$'s.
    Generically, adding this row will not decrease the Kruskal rank, i.e., $\krank(\vE[\vec Z' \mid U]) = I_1$, $\krank(\vE[\vec S' \mid U]) = I_2$, and $\krank(\vE[\vec S' \mid U]) = I_3$.
    
    Then, since $\vZ, \vX$, and $\vec S$ are conditionally independent given $U$, we have
    \[
    \vM[\vZ', \vX', \vec S']
    =
    [\vP(U) ; \vE[\vec Z' \mid U], \vE[\vec X' \mid U], \vE[\vec S' \mid U]]
    \]
    Under the assumption that $I_1 + I_2 + I_3 \geq 2k + 2$, the Kruskal rank theorem \citep{kruskal1977three} guarantees that the above rank-$k$ decomposition of $\vM[\vZ', \vX', \vec S']$ is unique up to permutation of the labels of $U$, and simultaneous scaling of the columns of each matrix.
    Under the additional constraint that the last rows of $\vE[\vec Z' \mid U]$, $\vE[\vec Z' \mid U]$, and $\vE[\vec Z' \mid U]$ are equal to one, the scaling indeterminacy is removed; i.e., the parameters are recoverable up to permutation.
\end{proof}

Given $\vP(U)$ and $\vE[\vec S \mid U]$, we can compute MTEs.
For example, we can compute the expected potential outcome under $T = 0$ for each subgroup $u$ as follows:
\begin{align*}
    \E [Y^{(0)} \mid U = u]
    &=
    \frac{\E (S_3 \mid U = u)}{\E (S_1 \mid U = u)}
    \\
    &=
    \frac{\E (Y \kron_{T=0} \mid U = u)}{\E(\kron_{T = 0} \mid U = u)}    
    \\
    &=
    \E (Y \mid U = u, T = 0),
\end{align*}
where the last line is a standard change of measure.

\subsection{ATE identifiability with moments (Level 4)}

Given a matrix $\vA$, we let $\vA^+$ denote the Moore-Penrose pseudoinverse of $\vA$.

\begin{lem}
    Let $\vX \indep \vZ \mid T, U$, let $Y \indep \vZ \mid T, U$, and let $\vX \indep T \mid U$.
    Assume that $\vM[\vec X, \vec Z \mid t]$ has a left inverse.
    Then
    \[
    \E[Y^{(t)}] = 
    \vM [Y, \vec Z \mid t]
    \cdot \vM[\vec X, \vec Z \mid t]^+ 
    \cdot \vE[\vec X]
    \]
\end{lem}
\begin{proof}
    First, for any $t$, since $\vX \indep \vZ \mid T, U$, we have by the law of total probability that
    \begin{equation*}
        \vM[\vec X, \vec Z \mid t]
        =
        \vE[\vec X \mid U, t]
        \cdot \diag(\vP(U \mid t))
        \cdot \vE[\vec Z \mid U, t]^\top
    \end{equation*}
    Similarly, for any $t$, since $Y \indep \vZ \mid T, U$, we have
    \begin{equation*}
        \vM[Y, \vec Z \mid t]
        =
        \vE[Y \mid U, t]
        \cdot \diag(\vP(U \mid t))
        \cdot \vE[\vec Z \mid U, t]^\top
    \end{equation*}
    Thus, if $\vM[\vec X, \vec Z \mid t]$ has a left inverse,
    \begin{equation}\label{eqn:exchange-moments-conditionals}
    \vM[Y, \vec Z \mid t]
    \cdot \vM[\vec X, \vec Z \mid t]^{+}
    =
    \vE[Y \mid U, t]
    \cdot \vE[\vec X \mid U, t]
    \end{equation}

    Now, we prove the result:
    \begin{align*}
        \E[Y^{(t)}]
        &=
        \vE[Y \mid U, t] 
        \cdot \vP[U]
        \\
        &=
        \vE[Y \mid U, t] 
        \cdot \vE[\vec X \mid U, t]^+ 
        \cdot \vE[\vec X \mid U, t]
        \cdot \vP[U]
        \tag{Multiply by the identity}
        \\
        &=
        \vM[Y, \vTheta] 
        \cdot \vM[\vec X, \vec Z]^+
        \cdot \vE[\vec X \mid U, t]
        \cdot \vP[U]
        \tag{Using \eqref{eqn:exchange-moments-conditionals}}
        \\
        &=
        \vM[Y, \vec Z] 
        \cdot \vM[\vec X, \vec Z]^+
        \cdot \vE[\vec X \mid U]
        \cdot \vP[U]
        \tag{Since $\vX \indep T \mid U$}
        \\
        &=
        \vM[Y, \vTheta] 
        \cdot \vM[\vec X, \vec Z]^+
        \cdot \vE[\vec X]
    \end{align*}
\end{proof}



\section{Additional experiments}

\begin{figure*}[!b]
    \centering
    \begin{subfigure}{0.33\textwidth}
    \includegraphics[width=\textwidth]{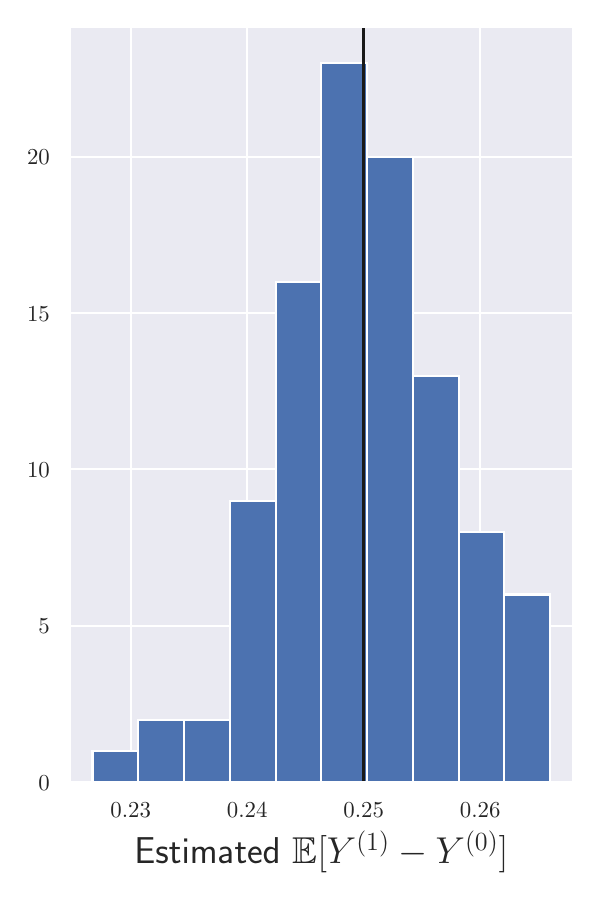}
    \caption{Experiment 1}
    \end{subfigure}
    \begin{subfigure}{0.66\textwidth}
    \includegraphics[width=\textwidth]{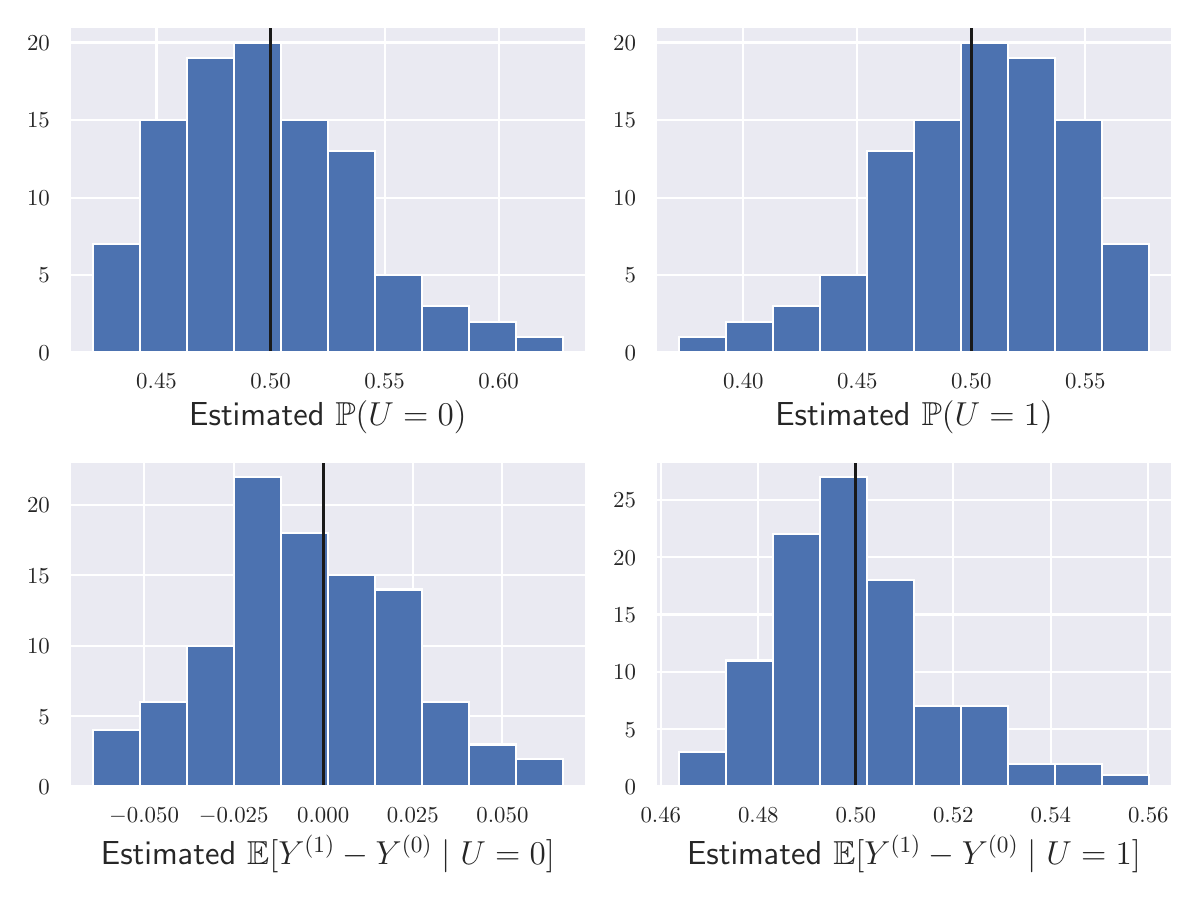}
    \caption{Experiment 2}
    \end{subfigure}
    \caption{
    \textbf{Synthetic Potential Outcomes accurately recover the ATE (average treatment effects), as well as the decomposition of the ATE into MTEs (mixed treatment effects)}. 
    In each plot, the true value is shown as a black vertical line, and the estimated values from 100 runs are shown as a histogram. See text for details.}
    \label{fig:experiments}
\end{figure*}
We have implemented our approach for computing SPOs and provide the following experiments on synthetic data. 
\paragraph{Experiment 1: ATEs}
We sample Bernoulli $U$ ($k=2$) uniformly as well as $4$ covariates $\abs{\vec{X}} = \abs{\vec{Z}} = 2$ from non-identical Bernoulli distributions that depend only on $U$.
For both experiments, we have the following causal mechanisms:
\begin{align*}
    \Pr(Z_1 = 1 \mid U) &= 0.2 + 0.3 \cdot U
    \\
    \Pr(Z_2 = 1 \mid U) &= 0.28 + 0.3 \cdot (1 - U)
    \\
    \Pr(X_1 = 1 \mid U) &= 0.36 + 0.3 \cdot U
    \\
    \Pr(X_2 = 1 \mid U) &= 0.44 + 0.3 \cdot (1 - U)
\end{align*}

These choices ensure that the matrices $\vec{M}[\vec Z, \vec X \mid T = 1]$ and $\vec{M}[\vec Z, \vec X \mid T = 0]$ in Algorithm~\ref{alg: SPOs} are relatively well-conditioned.
We sample $T, Y$ according to
\begin{equation}
    \begin{aligned}
        \Pr(T = 1)& = \frac{3}{4} - \frac{U}{2}\\
        \Pr(Y = 1) &= \frac{1}{4} + \frac{T}{4} + \frac{\kron_{U = T}}{4}.
    \end{aligned}
\end{equation}
We do this for $100$ runs of $100,000$ samples each and compute the difference between the true and calculated ATE in each run, reporting the results in Figure~\ref{fig:experiments}(a).
As expected, the method accurately estimates the average treatment effect $\mathbb{E}[Y^{(1)} - Y^{(0)}]$, despite unobserved confounding.

\paragraph{Experiment 2: MTEs} We sample from the same generative model as above, this time taking $100$ runs of $500,000$ samples each (for each parameter value) to account for the additional complexity of the task.
We report the results in Figure~\ref{fig:experiments}(b), showing that our method accurately decomposes the ATE into a mixture of two distinct underlying treatment effects.
Most importantly, the method recovers a clear separation between the two treatment effects, exposing the heterogeneity underlying the ATE that was recovered in Figure~\ref{fig:experiments}(a).

\end{document}